\pdfoutput=1

\documentclass[11pt]{article}

\usepackage[final]{acl}

\usepackage{times}
\usepackage{latexsym}
\usepackage{enumitem}

\usepackage[T1]{fontenc}

\usepackage[utf8]{inputenc}

\usepackage{microtype}

\usepackage{inconsolata}

\usepackage{graphicx}

\usepackage{times}
\usepackage{soul}
\usepackage{url}
\usepackage{hyperref}
\usepackage{caption}
\usepackage{subcaption}
\usepackage{graphicx}
\usepackage{amsmath}
\usepackage{amsthm}
\usepackage{booktabs}
\usepackage{algorithm}
\usepackage{algpseudocode}

\usepackage{multirow}
\usepackage{bbm}
\usepackage{mathrsfs}
\usepackage{amssymb,amsmath}

\usepackage{color}
\usepackage{xcolor}
\usepackage[switch]{lineno}
\usepackage{float}

\title{Position Specific Scoring Is All You Need? Revisiting Protein Sequence Classification Tasks}

\author{
 \textbf{Sarwan Ali\textsuperscript{1}},
 \textbf{Taslim Murad\textsuperscript{1}},
 \textbf{Prakash Chourasia\textsuperscript{1}}, \\
 \textbf{Haris Mansoor\textsuperscript{2}},
 \textbf{Imdad Ullah Khan\textsuperscript{2}},
 \textbf{Pin-Yu Chen\textsuperscript{3}},
 \textbf{Murray Patterson\textsuperscript{1}},
\\
\\
 \textsuperscript{1} Georgia State University, Atlanta, USA \\
 \textsuperscript{2} Lahore University of Management Sciences, Lahore, Pakistan \\
 \textsuperscript{3} IBM T. J. Watson Research Center, Yorktown Heights, Yorktown, NY, USA
}

\begin{document}
\maketitle
\begin{abstract}
Understanding the structural and functional characteristics of
proteins are crucial for developing preventative and curative
strategies that impact fields from drug discovery to policy
development. An important and popular technique for examining how
amino acids make up these characteristics of the protein sequences with position-specific
scoring (PSS). While the string kernel is crucial in natural language
processing (NLP), it is unclear if string kernels can extract
biologically meaningful information from protein sequences, despite
the fact that they have been shown to be effective in the general sequence
analysis tasks.  In this work, we propose a weighted PSS kernel matrix
(or W-PSSKM), that combines a PSS representation of protein sequences,
which encodes the frequency information of each amino acid in a sequence,
with the notion of the string kernel.  This results in a novel kernel
function that outperforms many other approaches for protein sequence
classification. 
We perform extensive experimentation to evaluate the proposed method. Our
findings demonstrate that the W-PSSKM significantly outperforms
existing baselines and state-of-the-art methods and achieves up to
45.1\% improvement in classification accuracy.
\end{abstract}


\section{Introduction}

Protein sequence classification is a fundamental problem in
bioinformatics, with applications in protein structure or function
prediction, viral host specificity, and drug
design~\cite{whisstock2003prediction,hirokawa1998sosui,rognan2007chemogenomic}. Various
feature-engineering-based methods to assist protein classification by
mapping sequences to numerical form are proposed, for example,
~\cite{kuzmin2020machine} creates one-hot encoding vectors.  Other
similar techniques~\cite{ali2021spike2vec,ma2020phylogenetic} work by
taking the $k$-mer frequencies and position distribution information
into account. Machine learning classifier uses these embeddings. However, the final prediction performance of such a pipeline may be affected by the particular properties associated with each embedding. For
example,
\cite{kuzmin2020machine} faces the curse of dimensionality challenge
and lacks model information regarding the relative positions of
amino acids. Similarly, \cite{ali2021spike2vec,ma2020phylogenetic} are
computationally expensive and can exhibit sparsity issues.

Several neural networks and transformer-based methods have been popularly used such as WDGRL~\cite{shen2018wasserstein}, AutoEncoder~\cite{xie2016unsupervised}, and Evolutionary Scale Modeling or ESM-2~\cite{lin2022language}. Additionally, several pre-trained models for protein classification, such as Protein Bert.~\cite{10.1093/bioinformatics/btac020}, Seqvec~\cite{heinzinger2019modeling}, UDSMProt~\cite{strodthoff2020udsmprot}, TAPE~\cite{rao2019evaluating} etc are proposed.

Another widely followed approach to protein sequence classification is
machine learning algorithms that make use of kernel functions,
particularly support vector machines (SVMs), due to their ability to
handle highly dimensional data and their robustness to
noise~\cite{xing2010brief}.  The kernel function --- a key component
of approaches such as SVM --- defines the similarity between pairs of
sequences.  For the classification of protein sequences, kernel-based
methods are typically favored over representation-based methods for
the following reasons:

\begin{itemize}[noitemsep]

\item
Protein sequences are complex, and contain structures (e.g., secondary or tertiary structure) that may not be
captured by the straightforward representation-based embeddings.
To handle this complexity, kernel-based techniques convert the sequences into a higher dimensional latent feature space. This feature space can be constructed by taking into account such complicated patterns or structures to classify the sequences~\cite{ali2022efficient}.


\item
Managing numerous forms of information: Protein sequences contain a
variety of data, including information about the secondary
structure, 
evolutionary relationships, and the makeup
composition of amino acids. By utilizing multiple kernel functions,
different sorts of information can be captured using kernel-based
methods.

\item
Better performance: It has been demonstrated that for protein sequence
classification, kernel-based algorithms outperform
representation-based methods in vasrios research
works~\cite{mismatchProteinClassification}
~\cite{counterMismatchKernel}. This is because
kernel-based approaches can successfully handle the complexity of the
sequences and collect more information, as mentioned above.

\end{itemize}

Although several efforts have been made to propose string kernel
methods in the
literature~\cite{ali2022efficient},
these methods are general-purpose, i.e., they are not designed to
consider the specific nature of protein sequences. To bridge this gap,
we propose a novel kernel function for protein sequence classification
called the weighted position-specific scoring kernel matrix
(W-PSSKM). 

Our contributions to this paper are the following:
\begin{enumerate}[noitemsep]

\item
We propose a kernel function, called W-PSSKM, to efficiently design a
kernel matrix specifically for protein sequences.

\item
Using kernel PCA, we design feature embeddings that enable the use of our
kernel matrix with a wide variety of non-kernel classifiers
(along with kernel classifiers, such as SVM) for supervised analysis
of protein sequences.

\item We provide a theoretical analysis of the W-PSSKM, including a proof of Mercer's theorem~\cite{mercer1909functions}.

\item
By demonstrating the performance of our kernel function on different
real-world protein sequence datasets, we show that the W-PSSKM
achieves high predictive accuracy and outperforms recent baselines and
state-of-the-art methods from the literature.

\end{enumerate}


\section{Related Work}\label{sec_related_work}
Biological sequence study is a popular topic in research, like protein analysis~\cite{buchan2019psipred} is essential for inferring its functional and structural properties, which helps in understanding diseases and building prevention mechanisms like drug discovery, etc.
Various feature embedding-based methods are put forward to gain a deeper understanding of the biological sequences like~\cite{kuzmin2020machine} proposed a one-hot encoding technique to classify spike protein sequences. PWKmer~\cite{ma2020phylogenetic} method uses position distribution information and $k$-mers frequencies to do a phylogenetic analysis of HIV-1 viruses. However, these methods are computationally expensive and can face the curse of dimensionality challenge. An extended version of the related work content is provided in Section~\ref{sec_related_work_extend} (in the appendix).

\section{Proposed Approach}\label{sec_proposed_approach}
In this section, we first discuss the algorithm and overall pipeline used to generate the proposed weighted kernel matrix (W-PSSKM). The theoretical proof of Mercer's theorem along with different properties of W-PSSKM is shown in Section~\ref{sec_theory}  (in the appendix).

\subsection{Proposed Algorithmn}
The generation of the Kernel matrix for protein sequences is the aim of this research work. The suggested method's algorithmic pseudocode is provided in Algorithms~\ref{algo_kernel} and~\ref{algo_PWM_fun}. Moreover, Figure~\ref{Weighted_PWM_flow_chart} (in the appendix) depicts the overall pipeline for our proposed method. There are three main steps involved in generating W-PSSKM: 

\paragraph{Step 1:} For the pair of given sequences ($seq_1$ and $seq_2$) and set of unique Amino Acids $L$ (where $L = 20$ for protein sequences), we first compute the position-specific scoring matrix (PSSM) using Algorithm~\ref{algo_PWM_fun}, which is called as a function in lines 6 and 7 of Algorithm~\ref{algo_kernel} (also shown in Figure~\ref{Weighted_PWM_flow_chart}-ii). 
In Algorithm~\ref{algo_PWM_fun}, we scan the given protein sequence and increment the respective row (position of amino acid in given sequence) and column (position of amino acid among $20$ unique amino acids i.e. ``ACDEFGHIKLMNPQRSTVWY'') value of position specific scoring matrix (performed using AAIndex function). Note that we use data padding in the scenario where sequences do not have fixed lengths. We noted that this step did not change the behavior of the supervised analysis in the case where we have unaligned sequence data.

\paragraph{Step 2:} After getting the position-specific scoring matrix (lines $6$ and $7$ of Algorithms~\ref{algo_kernel}) for two given sequences (i.e. $PSSM_1$ and $PSSM_2$), these matrices are used (column sum) to get the frequency count of each amino acid ($freq_1$, $freq_2$) for both sequences (see lines 8 and 9 in Algorithms~\ref{algo_kernel}). Then, the Weight vectors (i.e. $\Vec{w_1}$ and $\Vec{w_2}$) are computed by normalizing the frequencies as shown in lines $10$ and $11$ in  Algorithm~\ref{algo_kernel}.

\paragraph{Step 3:} Finally, the kernel value is computed by first taking the element-wise product of  $PSSM_1$ and $PSSM_2$ matrices and then multiplying with respective weight vectors (i.e. $\Vec{w_1}$ and $\Vec{w_2}$). This will give us the vectors $A$ and $B$ (see lines $12$ and $13$) in Algorithm~\ref{algo_kernel}.
The $A$ and $B$ vectors are added together and their sum is taken to get the resultant kernel Equation~\ref{eq:1}.
The complete flow of kernel value computation between two sequences is given in Figure~\ref{Weighted_PWM_flow_chart} (in the appendix),
Note that line 16 in Algorithm~\ref{algo_kernel} assigns the value to the lower half of the Kernel matrix since it is a symmetric matrix.

\begin{equation}
\scriptsize{
\label{eq:1}
\begin{aligned}
k(x_i,x_j)=\mathbbm{1}^T \{ {(PSSM_{i} \odot PSSM_{j})} w_{i} + \\
{(PSSM_{j} \odot PSSM_{i})} w_{j} \}
\end{aligned}
}
\end{equation}

Where $\mathbbm{1}$ is a column vector of ones used to perform summation. Suppose, 

\begin{equation} \label{eq:2}
\scriptsize
A={(PSSM_{1} \odot PSSM_{2})} w_{1}=[a_1, a_2, ... a_s]^T
 \end{equation}

\begin{equation} \label{eq:3}
\scriptsize
B={(PSSM_{2} \odot PSSM_{1})} w_{2}=[b_1, b_2, ... b_s]^T
 \end{equation}

Then,
\begin{equation} \label{eq:4}
\scriptsize
k(x_i,x_j)=\mathbbm{1}^T(A+B)=\sum_{i=1}^s  (a_i +b_i)
 \end{equation}

After generating the kernel matrix as shown in Figure~\ref{Weighted_PWM_flow_chart} (in the appendix) and Algorithm~\ref{algo_kernel}, we use kernel PCA~\cite{hoffmann2007kernel} to design the feature embeddings for protein sequences using the principal components. This enables us to use non-kernel classifiers (e.g. KNN) and compare them with the kernel classifiers (e.g. SVM) to perform supervised analysis on protein sequences. For SVM, we use W-PSSKM as input while for all other classifiers, we use kernel-PCA-based embeddings as input to perform supervised analysis.


In practice, a large data set leads to a large $K$, and storing $K$ may become a problem. Kernel-PCA can help in this regard to convert $K$ into a low-dimensional subspace. Since our data is highly non-linear, kernel PCA can find the non-linear manifold. Using the W-PSSKM kernel, the originally linear operations of PCA are performed in a reproducing kernel Hilbert space. It does so by mapping the data into a higher-dimensional space but then turns out to lie in a lower-dimensional subspace of it. So Kernel-PCA increases the dimensionality to be able to decrease it.
Using Kernel-PCA, we compute the top principal components from K and use them as embeddings for supervised analysis as input for any linear and nonlinear classifiers.

\begin{algorithm}[h!]
\caption{Weighted PSSM Kernel Matrix (W-PSSKM)}
\label{algo_kernel}
\begin{algorithmic}[1]
\scriptsize
\Statex \textbf{Input}: Set of Protein Sequences ($sequences$)
\Statex \textbf{Output}: Kernel Matrix ($K$)
\For{$ind_1 \in$ $1: \vert sequences \vert$}
\For{$ind_2 \in$ $ind_1: \vert sequences \vert$}
\If{$ind_1$ $\leq$ $ind_2$} \Comment{Only upper triangle}
\State $seq_1 \gets sequences[ind_1]$
\State $seq_2 \gets sequences[ind_2]$

\State $PSSM_1 \gets \Call{ComputePSSM}{seq_1}$ 
\State $PSSM_2 \gets \Call{ComputePSSM}{seq_2}$ 

\State $freq_1 \gets \Call{ColumnSum}{PSSM_1}$
\State $freq_2 \gets \Call{ColumnSum}{PSSM_2}$

\State $\Vec{w_1} \gets \frac{freq_1}{\text{sum}(freq_1)}$ \Comment{Compute weight vec.}
\State $\Vec{w_2} \gets \frac{freq_2}{\text{sum}(freq_2)}$ \Comment{Compute weight vec.}
\State $A \gets$ ($PSSM_1 \odot PSSM_2$)$\Vec{w_1}$
\State $B \gets$ $ (PSSM_2 \odot PSSM_1) \Vec{w_2}$
\State $\Vec{V} \gets A+B$  
\State K[$ind_1$, $ind_2$] $\gets \Call{Sum}{\Vec{V}}=\sum_{i=1}^s  (a_i +b_i)$
\State K[$ind_2$, $ind_1$] $\gets$ K[$ind_1$, $ind_2$] \Comment{Symmetry}
\EndIf
\EndFor
\EndFor
\State \Return K
\end{algorithmic}
\end{algorithm}

\begin{algorithm}[h!]
\caption{Compute PSSM Matrix}
\label{algo_PWM_fun}
\begin{algorithmic}[1]
\scriptsize
\Function{ComputePSSM}{$sequence$}
\State $s \gets \text{len}(sequence)$ \Comment{Compute length of sequence}
\State $L = 20$ \Comment{Unique Amino Acid Count}
\State $PSSM \gets \text{zeros}(s, L)$ \Comment{Initialize PSSM matrix}
\For{$i \gets 1$ \textbf{to} $s$} 
    \State $intAA \gets \Call{AAIndex}{sequence[i]}$
        \State $PSSM[i, intAA] ++$ 
\EndFor
\State \Return $PSSM$
\EndFunction
\end{algorithmic}
\end{algorithm}

\subsection{Theoretical Proof}
\label{sec_theory}
Mercer's theorem, as discussed by~\cite{xu2019generalized}, offers a fundamental underpinning for the utilization of kernel methods in machine learning. The theorem establishes that any positive semi-definite kernel function can be represented as the inner product of two functions in a feature space of high dimensionality. This enables us to bypass the explicit computation of the high-dimensional feature space and instead operate directly with the kernel function in the input space. Meeting the conditions of Mercer's theorem allows for the application of kernel methods like support vector machines (SVMs) and kernel principal component analysis (KPCA). These methods bring forth advantages such as efficient computation, the flexibility to model intricate relationships between data, and the ability to handle non-linearly separable data. Additionally, the desirable properties of positive definiteness and full rank, which will be discussed in detail below, offer further benefits such as facilitating the construction of a reproducible kernel Hilbert Space (RKHS). This framework provides a basis for a unique and reproducible Kernel.

A kernel function $k(x_i, x_j)\  \forall \ x_i,x_j \in \mathbb{R}^D$ is called a Mercer kernel if it satisfies the following properties:

\begin{enumerate}[noitemsep]
    \item $k(x_i, x_j)$ is symmetric, i.e. $k(x_i, x_j) = k(x_j, x_i)$ $\forall$ $x_i$ and $x_j$.
    \item For any set of data points ${x_1, x_2, ..., x_n}$, the kernel matrix $K$ defined by $K_{ij} = k(x_i, x_j)$ is positive semi-definite.
\end{enumerate}
\begingroup

\paragraph{\textbf{Symmetry: }}
The W-PSSKM kernel function is symmetric since, 
\begin{equation} \label{eq:5}
\scriptsize
k(x_i,x_j)=\sum_{i=1}^s  (a_i +b_i)=\sum_{i=1}^s  (b_i +a_i)=k(x_j,x_i)
 \end{equation}

\paragraph{\textbf{Positive semi-definiteness:}}
For any set of data points ${v_1, v_2, ..., v_n}$, the kernel matrix is defined as $K(i,j)=\sum_{i=1}^s  (a_i +b_i)$, is positive semi-definite if, $\{ v^TKv \geq 0, \forall \ v \in \mathbb{R}^{n} \}$. We can redefine  the kernel as $K(i,j)=\sum_{i=1}^s  (a_i +b_i)= \langle \phi(x_i)\phi(x_j) \rangle$ where, $\phi(x_i)=[\sqrt{a_1},\sqrt{b_1}, \sqrt{a_2},\sqrt{b_2},...\sqrt{a_s},\sqrt{b_s}]$.

 \begin{equation} \label{eq:6}
 \scriptsize
      v^TKv = \sum_{i=1}^n \sum_{j=1}^n v_i v_j K(i,j) 
 \end{equation}

 \begin{equation} \label{eq:7}
 \scriptsize
      v^TKv = \sum_{i=1}^n \sum_{j=1}^n v_i v_j \phi(x_i)\phi(x_j)
 \end{equation}

\begin{equation} \label{eq:8}
\scriptsize
 v^TKv =\left(\sum_{i=1}^n v_i \phi(x_i)\right) \left(\sum_{j=1}^n v_j \phi(x_j)\right)
 \end{equation}

 \begin{equation} \label{eq:9}
 \scriptsize
v^TKv =\left\| \left( \sum_{j=1}^n v_j \phi(x_j) \right) \right\|^{2} \geq 0
 \end{equation}

Since the weighted PSSKM kernel function is a Mercer kernel, there exists a function  $\phi$ that maps $x_i, x_j$ into another space (possibly with much higher dimensions) such that 
\begin{equation}\label{eq:10}
\scriptsize
k(x_i,x_j)=\phi(x_i)^T\phi(x_j)
\end{equation}
So you can use $K$ as a kernel since we know $\phi$ exists, even if you don’t know what $\phi$ is. We can use $\phi(x_i)^T\phi(x_j)$ as a comparison metric between samples. In a number of machine learning and statistical models, the feature vector $x$ only enters the model through comparisons with other feature vectors. If we kernelize these inner products, When we are allowed to compute only $k(x_i,x_j)$  in computation and avoid working in feature space, we work in an $n \times n$ space (the Kernel matrix).

\paragraph{\textbf{Positive Definite:}}
\textcolor{black}{Positive-definite kernels provide a framework that encompasses some basic Hilbert space constructions. A Hilbert space is called a reproducing kernel Hilbert space (RKHS) if the kernel functionals are continuous. Every reproducing kernel is positive-definite, and every positive definite kernel defines a unique RKHS, of which it is the unique reproducing kernel.} If Equation~\ref{eq:6} equates zero,
 \begin{equation} \label{eq:11}
 \scriptsize
      v^TKv = \sum_{i=1}^n \sum_{j=1}^n v_i v_j \sum_{k=1}^s(a_k,b_k)
 \end{equation}

Since both $a_k,b_k$ are positive for all $k$, the only way for Equation~\ref{eq:11} to be zero is when $v=0$, which is the condition of the positive definite matrix.  

\paragraph{\textbf{Full rank:}}
A matrix is a full rank if it has linearly independent columns. For the W-PSSKM kernel, the null space is given by $Kv =0$, multiplying $v^T$ on both sides, we get $v^TKv =0$
Since we have proved that $v^TKv =0$ is only possible for $v=0$ (Equation~\ref{eq:11}), the null space consists of only zero vector; thus, the matrix is full rank.

The runtime analysis for the proposed method is given in Section~\ref{sec_runtime}  (in the appendix).

\section{Experimental Evaluation}\label{sec_experimental_evaluation}
We use three datasets Spike7k, Protein Subcellular, and Coronavirus Host. The summary of all the datasets is given in Table~\ref{tbl_data_statistics} (in the appendix) and their detailed discussion can be found in Section~\ref{apen_dataset}  (in the appendix).
We use several baseline and state-of-the-art (SOTA) methods to compare performance with the proposed W-PSSKM kernel. These baseline and SOTA methods are summarized in Table~\ref{tbl_baselines} with detailed discussion in Section~\ref{apen_baselines}  (in the appendix).
Implementation is done in Python, pre-processed data and the code is available online for reproducibility~\footnote{available in the published version}. All experiments are conducted using an Intel(R) Xeon(R) CPU E7-4850 v4 @ 2.10GHz having Ubuntu 64-bit OS (16.04.7 LTS Xenial Xerus) with 3023 GB memory. Moreover, we used $70-30\%$ train-test data split with $10\%$ data from the training set used as a validation set for hyperparameters tuning. We repeat the experiments 5 times (on random splits) and report average and standard deviation (std.) results. The details of evaluation metrics and the ML classifiers used are given in Section~\ref{sec_eval_metric} (in the appendix).

\section{Results And Discussion}\label{sec_results}
In this section, we present the classification results for the proposed W-PSSKM and compare them with the results of baselines and SOTA methods on multiple datasets.

\paragraph{\textbf{Results for Coronavirus Host Dataset:}}
The average classification results (of $5$ runs) for Coronavirus Host data are reported in Table~\ref{tbl_results_host_classification}. We can observe that the proposed W-PSSKM outperforms baselines and SOTA methods for all but one evaluation metric (training runtime). On comparing the average accuracy, when compared to feature engineering-based techniques (Spike2Vec, PWM2Vec, Spaced $k$-mers), W-PSSKM shows up to $10.1 \%$ improvement in comparison to the second best (Spike2Vec with random forest and logistic regression). In comparison to NN-based models (WDGRL and Autoencoder), the W-PSSKM achieved up to $15.5 \%$ improvement to the second best (Autoencoder with random forest). 


In Table~\ref{tbl_results_host_classification}, compared to the string kernel method, which is designed while focusing on NLP problems in general, W-PSSKM achieves up to $28.7 \%$ improvement than the second best (string kernel with random forest). Moreover, while comparing to pre-trained language models for protein sequences (SeqVec and Protein Bert), the W-PSSKM achieves up to $12.2 \%$ improvement compared to the second best (SeqVec with random forest classifier). Overall, we can see that the proposed W-PSSKM significantly outperforms different types of baselines and SOTA methods from the literature. 
The standard deviation (std) results (of $5$ runs) for Coronavirus Host data are reported in Table~\ref{tbl_std_org_host}. With exception to the training runtime, the std values for all evaluation metrics are significantly lower for all evaluation metrics, baselines, SOTA, and the proposed W-PSSKM method. 
For training runtime, the WDGRL method with the Naive Bayes classifier takes the least time due to the fact that its embedding dimension is the lowest among others.
Compared to the end-to-end deep learning models (LSTM, GRU, and CNN), we can observe that the proposed approach significantly outperforms these models for all evaluation metrics. The main reason for the lower performance of deep learning models is that it is well established from previous works that deep learning (DL) methods do not work efficiently as compared to simple tree-based methods for tabular data~\cite{grinsztajn2022tree,joseph2022gate,malinin2020uncertainty}.

\begin{table}[h!]
    \centering
    \resizebox{0.45\textwidth}{!}{
    \begin{tabular}{p{1.6cm}lp{1.1cm}p{1.1cm}p{1.1cm}p{1.9cm}p{1.9cm}p{1.9cm}|p{1.6cm}}
    \toprule
        \multirow{2}{*}{Embeddings} & \multirow{2}{*}{Algo.} & \multirow{2}{*}{Acc. $\uparrow$} & \multirow{2}{*}{Prec. $\uparrow$} & \multirow{2}{*}{Recall $\uparrow$} & \multirow{2}{1.4cm}{F1 (Weig.) $\uparrow$} & \multirow{2}{1.5cm}{F1 (Macro) $\uparrow$} & \multirow{2}{1.2cm}{ROC AUC $\uparrow$} & Train Time (sec.) $\downarrow$ \\
        \midrule \midrule
        \multirow{7}{1.2cm}{Spike2Vec}
        & SVM & 0.848 & \underline{0.852} & 0.848 & 0.842 & 0.739 & \underline{0.883} & 191.066 	\\
        & NB & 0.661 & 0.768 & 0.661 & 0.661 & 0.522 & 0.764 & 10.220     \\
        & MLP & 0.815 & 0.837 & 0.815 & 0.814 & 0.640 & 0.835 & 46.624    \\
        & KNN & 0.782 & 0.794 & 0.782 & 0.781 & 0.686 & 0.832 & 82.112    \\
        & RF & \underline{0.853} & 0.848 & \underline{0.853} & 0.845 & 0.717 & 0.864 & 15.915     \\
        & LR & \underline{0.853} & \underline{0.852} & \underline{0.853} & \underline{0.846} & \underline{0.757} & 0.879 & 60.620     \\
        & DT & 0.829 & 0.827 & 0.829 & 0.825 & 0.696 & 0.855 & \underline{4.261}      \\
        
        \cmidrule{2-9}
        \multirow{7}{1.2cm}{PWM2Vec}
        & SVM & 0.799 & 0.806 & 0.799 & 0.801 & 0.648 & 0.859 & 44.793	\\
        & NB & 0.381 & 0.584 & 0.381 & 0.358 & 0.400 & 0.683 & \underline{2.494}    \\
        & MLP & 0.782 & 0.792 & 0.782 & 0.778 & 0.693 & 0.848 & 21.191  \\
        & KNN & 0.786 & 0.782 & 0.786 & 0.779 & 0.679 & 0.838 & 12.933  \\
        & RF & \underline{0.836} & \underline{0.839} & \underline{0.836} & \underline{0.828} & \underline{0.739} & \underline{0.862} & 7.690    \\
        & LR & 0.809 & 0.815 & 0.809 & 0.800 & 0.728 & 0.852 & 274.917  \\
        & DT & 0.801 & 0.802 & 0.801 & 0.797 & 0.633 & 0.829 & 4.537    \\

        \cmidrule{2-9}

        \multirow{7}{1.9cm}{Spaced $k$-mers}
        & SVM & 0.830 & 0.836 & 0.830 & 0.825 & 0.645 & 0.832 & 4708.264	\\
        & NB & 0.711 & 0.792 & 0.711 & 0.707 & 0.621 & 0.809 & 291.798      \\
        & MLP & 0.829 & 0.842 & 0.829 & 0.823 & 0.586 & 0.774 & 1655.708    \\
        & KNN & 0.780 & 0.783 & 0.780 & 0.775 & 0.589 & 0.790 & 2457.727    \\
        & RF & 0.842 & 0.850 & 0.842 & 0.835 & 0.632 & 0.824 & 542.910      \\
        & LR & \underline{0.844} & \underline{0.851} & \underline{0.844} & \underline{0.837} & \underline{0.691} & \underline{0.833} & 187.966      \\
        & DT & 0.830 & 0.839 & 0.830 & 0.826 & 0.640 & 0.827 & \underline{56.868}       \\
        
        \midrule

        \multirow{7}{1.9cm}{WDGRL}
                 & SVM & 0.329 & 0.108 & 0.329 & 0.163 & 0.029 & \underline{0.500} & 2.859 \\
         & NB & 0.004 & 0.095 & 0.004 & 0.007 & 0.002 &  0.496 & \textbf{0.008}  \\
         & MLP & 0.328 & 0.136 & 0.328 & 0.170 & 0.032 & 0.499 & 5.905  \\
         & KNN & 0.235 & 0.198 & 0.235 & 0.211 & \underline{0.058} & 0.499 & 0.081  \\
         & RF & 0.261 & 0.196 & 0.261 & \underline{0.216} & 0.051 & 0.499 &  1.288 \\
         & LR & \underline{0.332} & 0.149 & \underline{0.332} & 0.177 & 0.034 & \underline{0.500} & 0.365   \\
         & DT & 0.237 & \underline{0.202} & 0.237 & 0.211 & 0.054 & 0.498 & 0.026  \\ 
                 
         \cmidrule{2-9}
        \multirow{7}{1.5cm}{Auto-Encoder}
         & SVM & 0.602 & 0.588 & 0.602 & 0.590 &  0.519 & 0.759 & 2575.95 \\
         & NB & 0.261 & 0.520 & 0.261 & 0.303 & 0.294 & 0.673 & 21.747 \\
         & MLP & 0.486 & 0.459 & 0.486 & 0.458 & 0.216 &  0.594 & 29.933 \\
         & KNN & 0.763 & 0.764 & 0.763 & 0.755 & 0.547 & 0.784 & \underline{18.511} \\
         & RF &  \underline{0.800} & \underline{0.796} & \underline{0.800} & \underline{0.791} & \underline{0.648} & \underline{0.815} &  57.905 \\
         & LR & 0.717 & 0.750 & 0.717 & 0.702 & 0.564 & 0.812 & 11072.67 \\
         & DT & 0.772 & 0.767 & 0.772 & 0.765 & 0.571 & 0.808 & 121.362 \\ 
        \midrule
  \multirow{7}{1.6cm}{String Kernel}
    & SVM & 0.601 & 0.673 & 0.601 & 0.602 & 0.325 & 0.624 & 5.198 \\
     & NB & 0.230 & 0.665 & 0.230 & 0.295 & 0.162 & 0.625 & \underline{0.131} \\
     & MLP & 0.647 & 0.696 & 0.647 & 0.641 & 0.302 & 0.628 & 42.322 \\
     & KNN & 0.613 & 0.623 & 0.613 & 0.612 & 0.310 & 0.629 & 0.434 \\
     & RF & \underline{0.668} & 0.692 & \underline{0.668} & \underline{0.663} & \underline{0.360} & \underline{0.658} & 4.541 \\
     & LR & 0.554 & \underline{0.724} & 0.554 & 0.505 & 0.193 & 0.568 & 5.096 \\
     & DT &  0.646 & 0.674 & 0.646 & 0.643 & 0.345 & 0.653 & 1.561 \\
          \midrule
           \multirow{3}{1.6cm}{Neural Network}
          & LSTM & 0.325 & 0.103 & \underline{0.325} & 0.154 & 0.021 & 0.502 & 21634.34 \\
          & CNN & \underline{0.442} & 0.101 & 0.112 & 0.086 & \underline{0.071} & \underline{0.537} & \underline{17856.40} \\
          & GRU & 0.321 & \underline{0.139} & 0.321 & \underline{0.168} & 0.032 & 0.505 & 126585.01 \\
          \midrule
\multirow{7}{1.5cm}{SeqVec}
 & SVM & 0.711 & 0.745 & 0.711 & 0.698 & 0.497 & 0.747 & 0.751 \\
 & NB & 0.503 & 0.636 & 0.503 & 0.554 & 0.413 & 0.648 & \underline{0.012} \\
 & MLP & 0.718 & 0.748 & 0.718 & 0.708 & 0.407 & 0.706 & 10.191  \\
 & KNN & 0.815 & 0.806 & 0.815 & 0.809 & 0.588 & 0.800 & 0.418  \\
 & RF & \underline{0.833} & \underline{0.824} & \underline{0.833} & \underline{0.828} & \underline{0.678} & \underline{0.839} &  1.753   \\
 & LR & 0.673 & 0.683 & 0.673 & 0.654 & 0.332 & 0.660 & 1.177   \\
 & DT & 0.778 & 0.786 & 0.778 & 0.781 & 0.618 & 0.825 & 0.160  \\ 
           \cmidrule{2-9}
\multirow{1}{1.5cm}{Protein Bert}
 & \multirow{2}{*}{\_} & \multirow{2}{*}{0.799} &  \multirow{2}{*}{0.806} & \multirow{2}{*}{0.799} & \multirow{2}{*}{0.789} & \multirow{2}{*}{0.715} & \multirow{2}{*}{0.841} & \multirow{2}{*}{15742.95} \\
  &&&&&&& \\
 \cmidrule{2-9}
  \multirow{7}{1.6cm}{ESM-2}
  & SVM & 0.981 & 0.978 & 0.981 & 0.977 & 0.744 & \textbf{\underline{0.898}} & 8.463 \\
 & NB & 0.759 & 0.950 & 0.759 & 0.757 & 0.573 & 0.810 & 8.595 \\
 & MLP & 0.978 & 0.975 & 0.978 & 0.974 & 0.706 & 0.875 & 62.442 \\
 & KNN & 0.977 & 0.975 & 0.977 & 0.973 & 0.707 & 0.865 & \textbf{\underline{2.202}} \\
 & RF & \textbf{\underline{0.983}} & 0.978 & \textbf{\underline{0.983}} & \textbf{\underline{0.979}} & \textbf{\underline{0.783}} & 0.893 & 17.740 \\
 & LR & 0.982 & \textbf{\underline{0.979}} & 0.982 & 0.978 & 0.777 & 0.889 & 107.879 \\
 & DT & 0.982 & \textbf{\underline{0.979}} & 0.982 & 0.978 & 0.742 & 0.890 & 3.733 \\
\cmidrule{2-9}
  \multirow{7}{1.6cm}{TAPE}
  & SVM & 0.818 & 0.823 & 0.818 & 0.811 & 0.711 & \underline{0.854} & 3.201 \\
 & NB & 0.482 & 0.587 & 0.482 & 0.442 & 0.400 & 0.712 & 0.494 \\
 & MLP & 0.812 & 0.819 & 0.812 & 0.802 & 0.665 & 0.828 & 3.737 \\
 & KNN & 0.793 & 0.797 & 0.793 & 0.789 & 0.633 & 0.818 & \underline{0.150} \\
 & RF & \underline{0.830} & \underline{0.834} & \underline{0.830} & \underline{0.823} & \underline{0.725} & 0.846 & 13.656 \\
 & LR & 0.779 & 0.797 & 0.779 & 0.764 & 0.628 & 0.794 & 11.325 \\
 & DT & 0.785 & 0.786 & 0.785 & 0.782 & 0.578 & 0.798 & 4.675 \\

 \midrule
\multirow{7}{1.6cm}{W-PSSKM (ours)}
 & SVM & 0.952 & 0.950 & 0.952 & 0.950 & 0.704 & \underline{0.886} & 0.894 \\
 & NB & 0.634 & 0.710 & 0.634 & 0.634 & 0.426 & 0.725 & \underline{0.101} \\ 
 & MLP & 0.874 & 0.878 & 0.874 & 0.873 & 0.535 & 0.778 & 15.964 \\ 
 & KNN & 0.939 & 0.933 & 0.939 & 0.934 & 0.555 & 0.773 & 0.345 \\
 & RF & \underline{0.955} & \underline{0.953} & \underline{0.955} & \underline{0.951} & 0.710 & 0.833 & 4.933 \\
 & LR & 0.950 & 0.946 & 0.950 & 0.946 & \underline{0.775} & 0.847 & 13.007 \\ 
 & DT & 0.933 & 0.938 & 0.933 & 0.934 & 0.608 & 0.836 & 1.369 \\

         \bottomrule
         \end{tabular}
    }
    \caption{Classification results (averaged over $5$ runs) for different evaluation metrics for \textbf{Coronavirus Host Dataset}. The best values for each method are underlined, while the overall best values are shown in bold.}
    \label{tbl_results_host_classification}
\end{table}

\begin{table*}[h!]
\centering
\resizebox{0.99\textwidth}{!}{
 \begin{tabular}{@{\extracolsep{6pt}}p{1.5cm}lp{1.1cm}p{1.1cm}p{1.1cm}p{1.3cm}p{1.3cm}p{1.1cm}p{1.7cm}
 p{1.1cm}p{1.1cm}p{1.1cm}p{1.3cm}p{1.3cm}p{1.1cm}p{1.7cm}}
    \toprule
    & & \multicolumn{7}{c}{Spike7k} & \multicolumn{7}{c}{Protein Subcellular} \\
    \cmidrule{3-9} \cmidrule{10-16}
        \multirow{2}{*}{Embeddings} & \multirow{2}{*}{Algo.} & \multirow{2}{*}{Acc. $\uparrow$} & \multirow{2}{*}{Prec. $\uparrow$} & \multirow{2}{*}{Recall $\uparrow$} & \multirow{2}{1.7cm}{F1 (Weig.) $\uparrow$} & \multirow{2}{1.7cm}{F1 (Macro) $\uparrow$} & \multirow{2}{1.2cm}{ROC AUC $\uparrow$} & Train Time (sec.) $\downarrow$
          & \multirow{2}{*}{Acc. $\uparrow$} & \multirow{2}{*}{Prec. $\uparrow$} & \multirow{2}{*}{Recall $\uparrow$} & \multirow{2}{1.7cm}{F1 (Weig.) $\uparrow$} & \multirow{2}{1.7cm}{F1 (Macro) $\uparrow$} & \multirow{2}{1.2cm}{ROC AUC $\uparrow$} & Train Time (sec.) $\downarrow$\\
        \midrule \midrule
        \multirow{7}{1.2cm}{Spike2Vec}
         & SVM & 0.855 & 0.853 & 0.855 & 0.843 & 0.689 & \underline{0.843} & 61.112 & 0.445 & 0.453 & 0.445 & 0.445 & 0.362 & 0.648 & 19.054 \\
         & NB & 0.476 & 0.716 & 0.476 & 0.535 & 0.459 & 0.726 & 13.292  & 
         0.275 & 0.272 & 0.275 & 0.229 & 0.186 & 0.569 & \underline{0.328} \\         
         & MLP & 0.803 & 0.803 & 0.803 & 0.797 & 0.596 & 0.797 & 127.066  & 
         0.369 & 0.376 & 0.369 & 0.370 & 0.262 & 0.597 & 109.620 \\         
         & KNN & 0.812 & 0.815 & 0.812 & 0.805 & 0.608 & 0.794 & 15.970  & 
         0.333 & 0.440 & 0.333 & 0.303 & 0.208 & 0.560 & 0.868 \\         
         & RF & 0.856 & \underline{0.854} & 0.856 & \underline{0.844} & 0.683 & 0.839 & 21.141  & 
         0.435 & 0.409 & 0.435 & 0.365 & 0.193 & 0.575 & 23.502 \\         
         & LR & \underline{0.859} & 0.852 & \underline{0.859} & \underline{0.844} & \textbf{0.690} & 0.842 & 64.027  & \underline{0.494} & \underline{0.490} & \underline{0.494} & \underline{0.489} & \underline{0.391} & \underline{0.665} & 32.149 \\         
         & DT & 0.849 & 0.849 & 0.849 & 0.839 & 0.677 & 0.837 & \underline{4.286}  & 
         0.322 & 0.318 & 0.322 & 0.319 & 0.197 & 0.563 & 9.771 \\
        \cmidrule{2-9} \cmidrule{10-16}
        \multirow{7}{1.2cm}{PWM2Vec}
        & SVM & 0.818 & 0.820 & 0.818 & 0.810 & 0.606 & \underline{0.807} & 22.710  & 
        0.423 & 0.444 & 0.423 & 0.426 & 0.339 & 0.640 & 79.182 \\
        
         & NB & 0.610 & 0.667 & 0.610 & 0.607 & 0.218 & 0.631 & 1.456  & 
         0.293 & 0.312 & 0.293 & 0.241 & 0.206 & 0.581 & \underline{0.810} \\
         
         & MLP & 0.812 & 0.792 & 0.812 & 0.794 & 0.530 & 0.770 & 35.197  & 
         0.309 & 0.315 & 0.309 & 0.310 & 0.206 & 0.568 & 111.598 \\
         
         & KNN & 0.767 & 0.790 & 0.767 & 0.760 & 0.565 & 0.773 & \underline{1.033}  & 
         0.285 & 0.461 & 0.285 & 0.247 & 0.192 & 0.549 & 1.964 \\
         
         & RF & \underline{0.824} & \underline{0.843} & \underline{0.824} & \underline{0.813} & \underline{0.616} & 0.803 & 8.290  & 
         0.436 & 0.496 & 0.436 & 0.379 & 0.210 & 0.577 & 84.261 \\
         
         & LR & 0.822 & 0.813 & 0.822 & 0.811 & 0.605 & 0.802 & 471.659  & 
         \underline{0.470} & \underline{0.476} & \underline{0.470} & \underline{0.470} & \underline{0.351} & \underline{0.645} & 96.467 \\
         
         & DT & 0.803 & 0.800 & 0.803 & 0.795 & 0.581 & 0.791 & 4.100  & 
         0.306 & 0.316 & 0.306 & 0.310 & 0.196 & 0.561 & 34.803 \\
         \cmidrule{2-9} \cmidrule{10-16}
 \multirow{7}{1.9cm}{Spaced $k$-mers} 
            & SVM & 0.852 & 0.841 & 0.852 & 0.836 & 0.678 & 0.840 & 2218.347  & 
            0.433 & 0.480 & 0.433 & 0.445 & 0.362 & 0.675 & 18.694 \\
            
            & NB & 0.655 & 0.742 & 0.655 & 0.658 & 0.481 & 0.749 & 267.243  & 
            0.266 & 0.304 & 0.266 & 0.207 & 0.197 & 0.582 & 1.087 \\
            
            & MLP & 0.809 & 0.810 & 0.809 & 0.802 & 0.608 & 0.812 & 2072.029  & 
            0.437 & 0.445 & 0.437 & 0.440 & 0.307 & 0.624 & 165.334 \\
            
            & KNN & 0.821 & 0.810 & 0.821 & 0.805 & 0.591 & 0.788 & \underline{55.140}  & 
            0.311 & 0.493 & 0.311 & 0.291 & 0.243 & 0.568 & \underline{0.734} \\
            
            & RF & 0.851 & 0.842 & 0.851 & 0.834 & 0.665 & 0.833 & 646.557  & 
            0.465 & \underline{0.515} & 0.465 & 0.415 & 0.241 & 0.591 & 33.646 \\
            
            & LR & \underline{0.855} & 0.848 & \underline{0.855} & 0.840 & 0.682 & 0.840 & 200.477  & 
            \underline{0.504} & 0.506 & \underline{0.504} & \underline{0.503} & \underline{0.407} & \underline{0.678} & 44.463 \\
            
            & DT & 0.853 & \underline{0.850} & 0.853 & \underline{0.841} & \underline{0.685} & \underline{0.842} & 98.089  & 
            0.299 & 0.306 & 0.299 & 0.302 & 0.197 & 0.562 & 9.192 \\
            \midrule
       
           \multirow{7}{1.2cm}{WDGRL}  & SVM & 0.792 & 0.769 & 0.792 & 0.772 & 0.455 & 0.736 & 0.335 & 
           \underline{0.229} & 0.098 & \underline{0.229} & 0.137 & 0.057 & 0.503 & 1.752 \\
           
             & NB & 0.724 & 0.755 & 0.724 & 0.726 & 0.434 & 0.727 & 0.018  & 
             0.206 & 0.154 & 0.206 & 0.158 & 0.073 & 0.501 & \textbf{0.008} \\
             
             & MLP & 0.799 & 0.779 & 0.799 & 0.784 & 0.505 & 0.755 & 7.348  & 
             0.218 & 0.136 & 0.218 & 0.151 & 0.067 & 0.502 & 11.287 \\
             
             & KNN & \underline{0.800} & \underline{0.799} & \underline{0.800} & \underline{0.792} & 0.546 & 0.766 & 0.094  & 
             0.170 & 0.154 & 0.170 & 0.158 & \underline{0.086} & 0.500 & 0.273 \\
             
             & RF & 0.796 & 0.793 & 0.796 & 0.789 & \underline{0.560} & \underline{0.776} & 0.393  & 
             0.211 & \underline{0.167} & 0.211 & \underline{0.163} & 0.079 & \underline{0.503} & 2.097 \\
             
             & LR & 0.752 & 0.693 & 0.752 & 0.716 & 0.262 & 0.648 & 0.091  & 
              \underline{0.229} & 0.098 & \underline{0.229} & 0.137 & 0.057 & \underline{0.503} & 0.112 \\
              
             & DT & 0.790 & \underline{0.799} & 0.790 & 0.788 & 0.557 & 0.768 & \textbf{0.009}  & 
             0.152 & 0.154 & 0.152 & 0.153 & \underline{0.086} & 0.498 & 0.082 \\
 
  \cmidrule{2-9} \cmidrule{10-16}
\multirow{7}{1.5cm}{Auto-Encoder}
 & SVM &  0.699 & 0.720 & 0.699 & 0.678 & 0.243 & 0.627 & 4018.028  & 
 0.431 & 0.447 & 0.431 & 0.435 & 0.315 & 0.632 & 95.840 \\
 
 & NB & 0.490 & 0.533 & 0.490 & 0.481 & 0.123 & 0.620 & 24.6372  & 
 0.228 & 0.305 & 0.228 & 0.205 & 0.161 & 0.569 & \underline{0.316} \\
 
 & MLP & 0.663 & 0.633 & 0.663 & 0.632 & 0.161 & 0.589 & 87.4913  & 
 0.412 & 0.389 & 0.412 & 0.399 & 0.253 & 0.598 & 126.795 \\
 
 & KNN & 0.782 & 0.791 & 0.782 &  0.776 & 0.535 & 0.761 & \underline{24.5597}  & 
 0.275 & 0.292 & 0.275 & 0.219 & 0.127 & 0.529 & 1.970 \\
 
 & RF & \underline{0.814} & \underline{0.803} & \underline{0.814} & \underline{0.802} & \underline{0.593} & \underline{0.793} &  46.583  & 
 0.381 & 0.347 & 0.381 & 0.306 & 0.163 & 0.558 & 30.260 \\
 
 & LR & 0.761 & 0.755 & 0.761 & 0.735 & 0.408 & 0.705 & 11769.02  & 
 \underline{0.464} & \underline{0.452} & \underline{0.464} & \underline{0.455} & \underline{0.332} & \underline{0.639} & 138.959 \\
 
 & DT & 0.803 & 0.792 & 0.803 & 0.792 & 0.546 & 0.779 & 102.185  & 
 0.228 & 0.232 & 0.228 & 0.229 & 0.150 & 0.533 & 15.367 \\
\midrule
 \multirow{7}{1.9cm}{String Kernel}
        & SVM  & 0.845 & 0.833 & \underline{0.846} & 0.821 & 0.631 & 0.812 & 7.350  & 
         0.496 & 0.510 & 0.496 & 0.501 & 0.395 & 0.674 & 5.277 \\
         
         & NB   & 0.753 & 0.821 & 0.755 & 0.774 & 0.602 & 0.825 & \underline{0.178}  & 
         0.301 & 0.322 & 0.301 & 0.265 & 0.243 & 0.593 & \underline{0.136} \\
         
         & MLP  & 0.831 & 0.829 & 0.838 & 0.823 & 0.624 & 0.818 & 12.652  & 
         0.389 & 0.390 & 0.389 & 0.388 & 0.246 & 0.591 & 7.263 \\
         
         & KNN  & 0.829 & 0.822 & 0.827 & 0.827 & 0.623 & 0.791 & 0.326  & 
         0.372 & 0.475 & 0.372 & 0.370 & 0.272 & 0.586 & 0.395 \\
         
         & RF   & \underline{0.847} & \underline{0.844} & 0.841 & \underline{0.835} & \underline{0.666} & 0.824 & 1.464  & 
         0.473 & 0.497 & 0.473 & 0.411 & 0.218 & 0.585 & 7.170 \\
         
         & LR   & 0.845 & 0.843 & 0.843 & 0.826 & 0.628 & 0.812 & 1.869  & 
         \underline{0.528} & \underline{0.525} & \underline{0.528} & \underline{0.525} & \underline{0.415} & \underline{0.678} & 8.194 \\
         
         & DT   & 0.822 & 0.829 & 0.824 & 0.829 & 0.631 & \underline{0.826} & 0.243  & 
         0.328 & 0.335 & 0.328 & 0.331 & 0.207 & 0.568 & 2.250 \\

         \midrule
           \multirow{3}{1.6cm}{Neural Network}
          & LSTM & 0.477 & 0.228 & 0.477 & 0.308 & 0.029 & 0.500 & 29872.920 & 0.442 & 0.446 & 0.442 & 0.439 & 0.441 & 0.512 & 35246.24 \\
          & CNN & 0.138 & 0.028 & 0.0719 & 0.0411 & 0.0232 &  \underline{0.503} &  \underline{2902.425} & 
          \underline{0.458} & \underline{0.459} & \underline{0.458} & 0.442 & 0.449 & \underline{0.537} & \underline{4157.84} \\
          & GRU & \underline{0.498} & \underline{0.248} & \underline{0.498} & \underline{0.331} & \underline{0.030} & 0.500 & 16191.921 & 
          0.451 & 0.449 & 0.451 & \underline{0.450} & \underline{0.458} & 0.512 & 34128.73 \\
            \midrule
        \multirow{7}{1.5cm}{SeqVec}
         & SVM & \underline{0.796} & 0.768 & \underline{0.796} & 0.770 & 0.479 & 0.747 & 1.0996  & 
         0.412 & 0.425 & 0.412 & \underline{0.421} & 0.306 & 0.611 & 10.241 \\
         
         & NB & 0.686 & 0.703 & 0.686 & 0.686 & 0.351 & 0.694 & \underline{0.0146}  & 
         0.205 & 0.297 & 0.205 & 0.196 & 0.154 & 0.542 & \underline{0.125} \\
         
         & MLP & \underline{0.796} & 0.771 & \underline{0.796} & 0.771 & 0.510 & 0.762 & 13.172  & 
         0.403 & 0.377 & 0.404 & 0.384 & 0.231 & 0.574 & 21.495 \\
         
         & KNN & 0.790 & 0.787 & 0.790 & \underline{0.786} & \underline{0.561} & 0.768 &  0.6463  & 
         0.244 & 0.271 & 0.245 & 0.201 & 0.114 & 0.511 & 1.141 \\
         
         & RF & 0.793 & \underline{0.788} & 0.793 & \underline{0.786} & 0.557 & \underline{0.769} &  1.8241  & 
         0.362 & 0.323 & 0.362 & 0.295 & 0.155 & 0.541 & 5.137 \\
         
         & LR & 0.785 & 0.763 & 0.785 & 0.761 & 0.459 & 0.740 & 1.7535  & 
         \underline{0.451} & \underline{0.444} & \underline{0.451} & \underline{0.421} & \underline{0.323} & \underline{0.624} & 4.427 \\
         
         & DT & 0.757 & 0.756 & 0.757 & 0.755 & 0.521 & 0.760 & 0.1308   & 
         0.213 & 0.221 & 0.213 & 0.224 & 0.149 & 0.517 & 7.752 \\
         \cmidrule{2-9} \cmidrule{10-16}
        \multirow{1}{1.9cm}{Protein Bert}
         & \_ &  0.836 & 0.828 & 0.836 & 0.814 & 0.570 & 0.792 & 14163.52  &  
          0.718 & 0.715 & 0.718 & 0.706 & 0.572 & 0.765 & 16341.85 \\
         \cmidrule{2-9} \cmidrule{10-16}
          \multirow{7}{1.5cm}{ESM-2} 
         & SVM & \underline{0.480} & \underline{0.231} & \underline{0.480} & \underline{0.312} & \underline{0.029} & \underline{0.500} & 15.501 & 
         0.868 & 0.872 & 0.868 & 0.869 & 0.840 & 0.915 & 250.665 \\
         & NB & 0.007 & 0.002 & 0.007 & 0.001 & 0.003 & 0.501 & 1.114 & 
         0.748 & 0.872 & 0.748 & 0.734 & 0.767 & 0.894 & 12.225 \\
         & MLP & \underline{0.480} & \underline{0.231} & \underline{0.480} & \underline{0.312} & \underline{0.029} & \underline{0.500} & 1.517 & 
         0.874 & 0.875 & 0.874 & 0.874 & 0.818 & 0.901 & 129.484 \\         
         & KNN & \underline{0.480} & \underline{0.231} & \underline{0.480} & \underline{0.312} & \underline{0.029} & \underline{0.500} & 0.426 & 
         0.845 & 0.859 & 0.845 & 0.841 & 0.758 & 0.871 & \underline{6.064} \\
         & RF & \underline{0.480} & \underline{0.231} & \underline{0.480} & \underline{0.312} & \underline{0.029} & \underline{0.500} & 3.208 & 
         \underline{0.911} & \underline{0.906} & \underline{0.911} & \underline{0.898} & 0.851 & 0.920 & 20.011 \\
         & LR & \underline{0.480} & \underline{0.231} & \underline{0.480} & \underline{0.312} & \underline{0.029} & \underline{0.500} & 8.527 & 
         0.898 & 0.890 & 0.898 & 0.894 & \textbf{\underline{0.855}} & \textbf{\underline{0.923}} & 267.809 \\
         & DT & \underline{0.480} & \underline{0.231} & \underline{0.480} & \underline{0.312} & \underline{0.029} & \underline{0.500} & \underline{0.117} & 
         0.841 & 0.846 & 0.841 & 0.843 & 0.814 & 0.900 & 33.622 \\
        \cmidrule{2-9} \cmidrule{10-16}
          \multirow{7}{1.5cm}{TAPE} 
         & SVM & \underline{0.831} & \underline{0.823} & \underline{0.831} & \underline{0.820} & \underline{0.628} & \underline{0.810} & 3.202 & 0.637 & 0.640 & 0.637 & 0.636 & 0.552 & \underline{0.760} & 8.553 \\
         & NB & 0.467 & 0.593 & 0.467 & 0.481 & 0.116 & 0.587 & 0.701 & 0.377 & 0.508 & 0.377 & 0.375 & 0.300 & 0.662 & \underline{0.311} \\
         & MLP & 0.813 & 0.794 & 0.813 & 0.795 & 0.542 & 0.775 & 6.304 & 0.590 & 0.590 & 0.590 & 0.589 & 0.432 & 0.695 & 5.296 \\
         & KNN & 0.783 & 0.783 & 0.783 & 0.774 & 0.556 & 0.765 & \underline{0.222} & 0.595 & 0.600 & 0.595 & 0.589 & 0.468 & 0.710 & 0.160 \\
         & RF & 0.813 & 0.805 & 0.813 & 0.799 & 0.611 & 0.787 & 13.760 & 0.600 & 0.622 & 0.600 & 0.572 & 0.405 & 0.666 & 28.819 \\ 
         & LR & 0.773 & 0.733 & 0.773 & 0.738 & 0.323 & 0.662 & 16.247 & \underline{0.671} & \underline{0.664} & \underline{0.671} & \underline{0.660} & \underline{0.553} & 0.746 & 15.968 \\
         & DT & 0.784 & 0.782 & 0.784 & 0.777 & 0.552 & 0.775 & 4.772 & 0.417 & 0.424 & 0.417 & 0.420 & 0.300 & 0.620 & 11.233 \\

\midrule

\multirow{7}{1.9cm}{W-PSSKM (ours)}
 & SVM & \textbf{\underline{0.861}} & \textbf{\underline{0.855}} & \textbf{\underline{0.861}} & \textbf{\underline{0.853}} & 0.594 & \textbf{\underline{0.844}} & 4.325  & 
 0.665 & 0.702 & 0.665 & 0.668 & 0.548 & 0.771 & 11.165 \\
 & NB & 0.549 & 0.610 & 0.549 & 0.522 & 0.374 & 0.687 & \underline{0.257}  & 
 0.308 & 0.483 & 0.308 & 0.311 & 0.291 & 0.646 & 0.166 \\ 
 & MLP & 0.741 & 0.746 & 0.741 & 0.743 & 0.529 & 0.756 & 45.670  & 
 0.729 & 0.731 & 0.729 & 0.727 & 0.617 & 0.798 & 3.096 \\ 
 & KNN & 0.734 & 0.750 & 0.734 & 0.733 & 0.375 & 0.669 & 1.024  & 
 0.909 & 0.913 & 0.909 & \textbf{\underline{0.909}} & \underline{0.803} & \underline{0.898} & \underline{0.117} \\
 & RF & 0.790 & 0.776 & 0.790 & 0.772 & 0.543 & 0.752 & 18.655  & 
 \textbf{\underline{0.915}} & \textbf{\underline{0.918}} & \textbf{\underline{0.915}} & 0.907 & 0.789 & 0.865 & 18.523 \\
 & LR & 0.837 & 0.824 & 0.837 & 0.818 & \underline{0.598} & 0.791 & 30.758 & 
 0.380 & 0.388 & 0.380 & 0.337 & 0.236 & 0.578 & 15.438 \\ 
 & DT & 0.751 & 0.758 & 0.751 & 0.753 & 0.503 & 0.750 & 7.443  & 
 0.855 & 0.856 & 0.855 & 0.854 & 0.758 & 0.871 & 3.435 \\ 
         \cmidrule{1-9} \cmidrule{10-16}
         \end{tabular}
}
 \caption{Classification results (averaged over $5$ runs) on \textbf{Spike7k} and \textbf{Protein Subcellular} datasets for different evaluation metrics. The best values for each method are underlined, while overall
best values are shown in bold.
 }
    \label{tbl_results_classification_7000}
\end{table*}

\paragraph{Results For Spike7K and Protein Subcellular}
\label{result_spike7k_protein}
The average classification results (of $5$ runs) for Spike7k and Protein subcellular data are reported in Table~\ref{tbl_results_classification_7000}.  
We can again observe that the proposed W-PSSKM outperforms all baseline and SOTA methods for the majority of the evaluation metrics. 
For the Spike7k dataset, we can observe that although the performance improvement of the proposed W-PSSKM is very small compared to all baselines and SOTA methods (for average accuracy, precision, recall, weighted F1, and ROC-AUC), it is still able to show its effectiveness by offering best performance overall.
For protein subcellular data, the proposed W-PSSKM achieves up to $41.1 \%$ improvement in average accuracy compared to the second-best feature engineering-based method (Spaced $k$-mers with logistic regression classifier). Similarly, the W-PSSKM achieves up to $45.1 \%$ improvement in average accuracy compared to the second-best NN-based method (Autoencoder with logistic regression classifier). 
Compared to the String kernel method, the proposed W-PSSKM achieves up to $38.7 \%$ improvement in average accuracy (compared to String kernel with logistic regression classifier). 
Finally, compared to pre-trained SOTA methods (SecVec and Protein Bert), the proposed W-PSSKM achieves up to $19.7 \%$ improvement in average accuracy (while comparing with the second best i.e. Protein Bert approach). 

The std results (of $5$ runs) for Spike7k and Protein Subcellular data are reported in Table~\ref{tbl_results_classification_std} (in the appendix). We can again observe that except for the training runtime, the std values for all evaluation metrics are significantly lower for all evaluation metrics, baselines, SOTA, and the proposed W-PSSKM method. Compared to the end-to-end deep learning models, we can again observe that the proposed method significantly outperforms these methods for all evaluation metrics in terms of predictive performance. 
We computed the confusion matrices to show the class-wise performance and highlighted the classes for which the method fails. The confusion matrix for the ESM-2-based model and our proposed method on the Protein Subcellular dataset using the Random Forest classifier is shown in Table ~\ref{Table_Confusion_Matrix_Prot_Sub_ESM-2} and ~\ref{Table_Confusion_Matrix_Prot_Sub_W_PWM} (in the appendix). 



\subsection{Class-wise Comparison}
\label{sec_class_wise_appendix}
An example of a pair of proteins belonging to the same
class and, to a different class
is shown in Figure~\ref{fig_barplot_5558} for the Coronavirus Host dataset. Figure~\ref{fig_barplot_5558} (a) and (b) represent the k-
mers spectrum for the environment label. As they belong to the
same class, we expect pairwise distance to be small and a large kernel value. The Euclidean distance metric cannot capture the
similarity effectively as compared to the W-PSSKM-based distance
measure. Similarly, Figure~\ref{fig_barplot_5558} (c) and (d) represent k-mers for
different classes, and we expect the Euclidean measure to
return large values and kernel to smaller values. However, W-PSSKM distance can capture
these differences more effectively than simple Euclidean-based
distance metrics. A similar example for Spike 7K and the Protein Subcellular dataset can be observed in Figure~\ref{fig_barplot_7000} and~\ref{fig_barplot_5959}, respectively.

\begin{figure}[h!]
\begin{subfigure}{.24\textwidth}
  \centering
  \includegraphics[scale = 0.180] {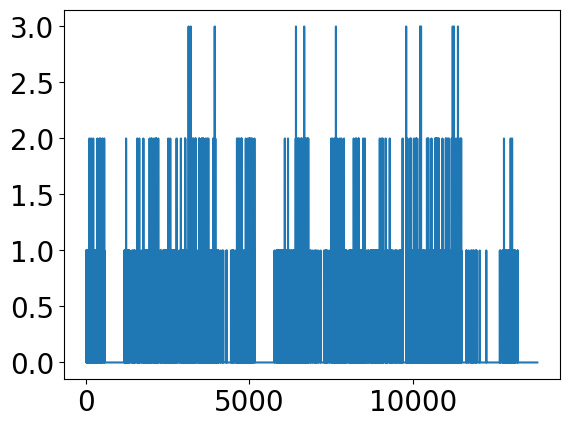}
  \caption{ Environment}
\end{subfigure}%
\begin{subfigure}{.24\textwidth}
  \centering
  \includegraphics[scale = 0.180] {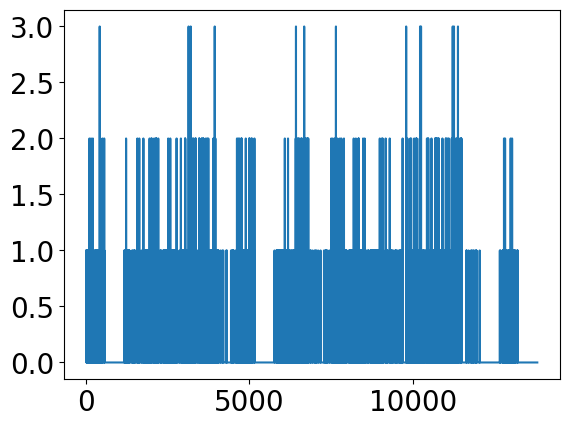}
  \caption{Environment}
\end{subfigure}%
\\
\begin{subfigure}{.24\textwidth}
  \centering
  \includegraphics[scale = 0.180] {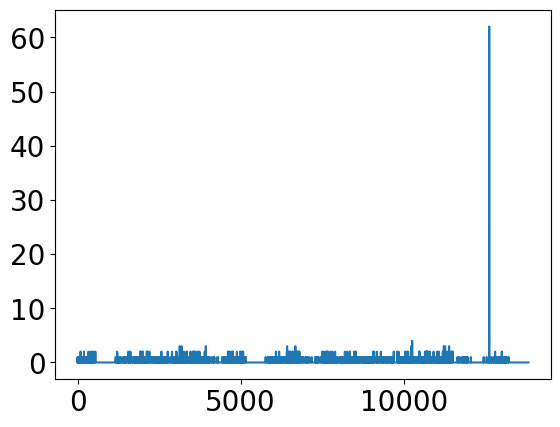}
  \caption{Cat}
\end{subfigure}%
\begin{subfigure}{.24\textwidth}
  \centering
  \includegraphics[scale = 0.180] {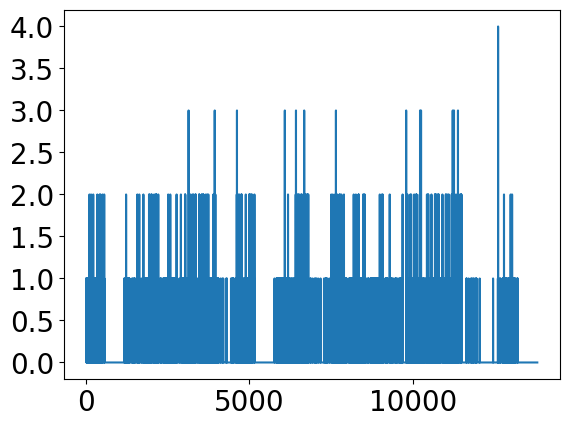}
  \caption{Weasel}
\end{subfigure}

\caption{K-mers spectrum of two pairs of classes. (a) and (b) belongs to the same class, while (c) and (d) belong to different classes for \textbf{Coronavirus Host dataset}. The Gaussian kernel distance for (a) and (b) is almost 0 while
for the W-PSSKM model is \textbf{3.23} (larger distance is better). The Gaussian kernel for (c) and (d) is \textbf{0.48} while for the W-PSSKM model is \textbf{0.39}  (smaller distance is better).}
 \label{fig_barplot_5558}
\end{figure}

\begin{figure}[h!]
\begin{subfigure}{.24\textwidth}
  \centering
  \includegraphics[scale = 0.180] {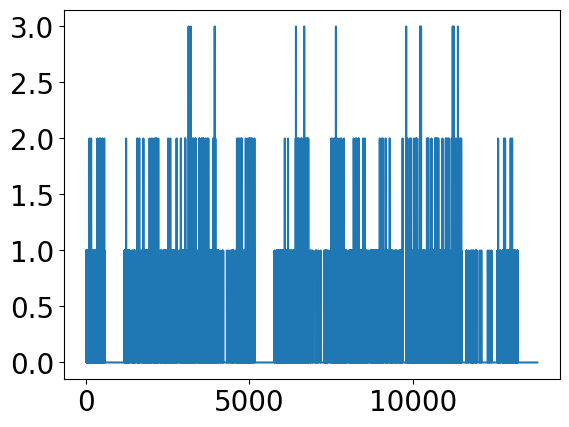}
  \caption{ B.1.1.7}
\end{subfigure}%
\begin{subfigure}{.24\textwidth}
  \centering
  \includegraphics[scale = 0.180] {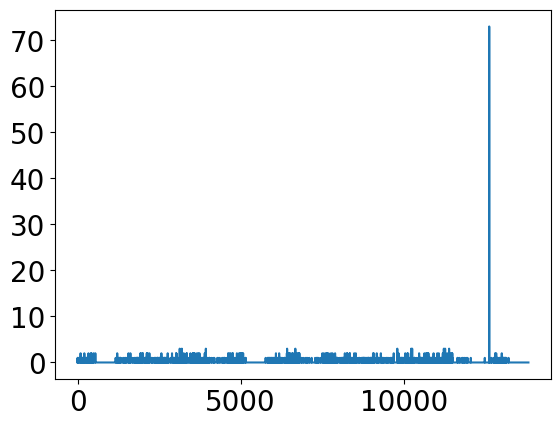}
  \caption{B.1.1.7}
\end{subfigure}%
\\
\begin{subfigure}{.24\textwidth}
  \centering
  \includegraphics[scale = 0.180] {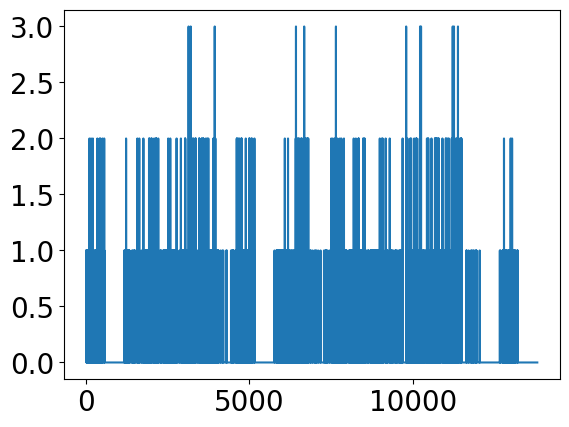}
  \caption{D.2}
\end{subfigure}%
\begin{subfigure}{.24\textwidth}
  \centering
  \includegraphics[scale = 0.180] {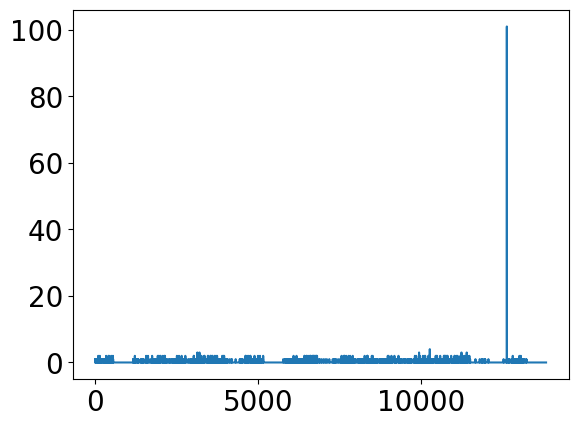}
  \caption{B.1.617.2}
\end{subfigure}

\caption{K-mers spectrum of two pairs of classes. (a) and (b) belongs to the same class, while (c) and (d) belong to different classes for \textbf{Spike7K dataset}. The Gaussian kernel distance for (a) and (b) is almost 0 while
for W-PSSKM model is \textbf{4.4} (larger distance is better). The Gaussian kernel for (c) and (d) is \textbf{0.57} while for W-PSSKM model is \textbf{0.49}  (smaller distance is better).}
 \label{fig_barplot_7000}
\end{figure}

\begin{figure}[h!]
\begin{subfigure}{.24\textwidth}
  \centering
  \includegraphics[scale = 0.180] {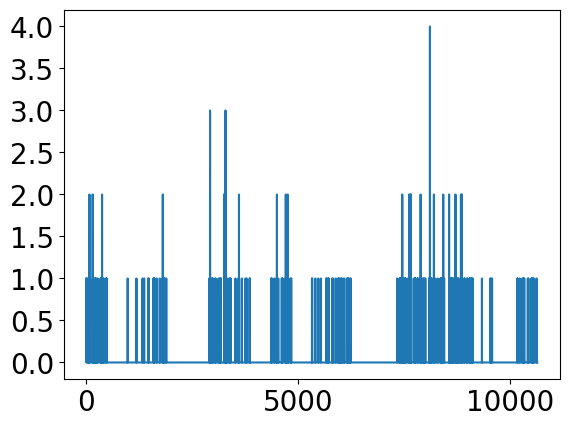}
  \caption{Chloroplast}
\end{subfigure}%
\begin{subfigure}{.24\textwidth}
  \centering
  \includegraphics[scale = 0.180] {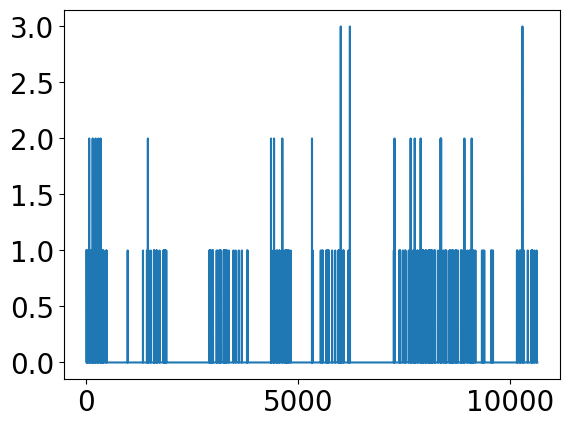}
  \caption{Chloroplast}
\end{subfigure}%
\\
\begin{subfigure}{.24\textwidth}
  \centering
  \includegraphics[scale = 0.180] {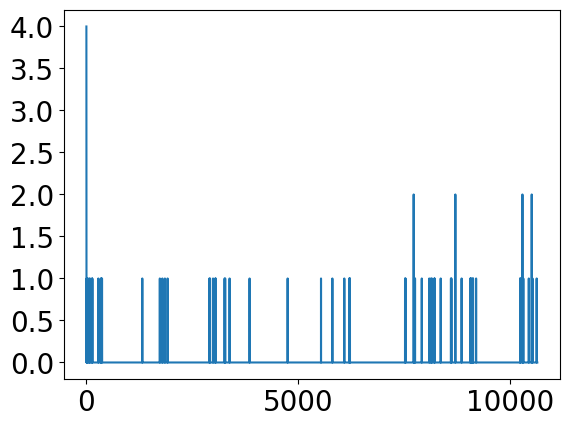}
  \caption{ER}
\end{subfigure}%
\begin{subfigure}{.24\textwidth}
  \centering
  \includegraphics[scale = 0.180] {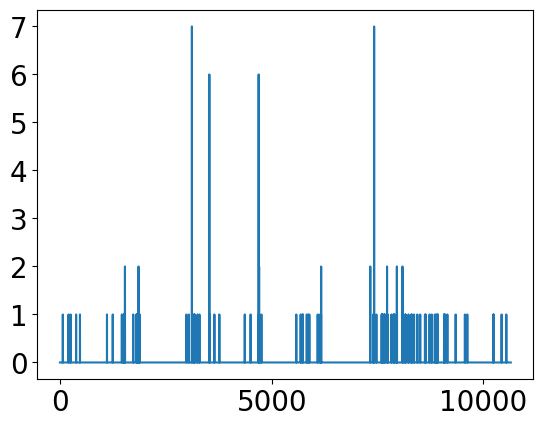}
  \caption{Extracellular}
\end{subfigure}
\caption{K-mers spectrum of two pairs of classes. (a) and (b) belongs to the same class, while (c) and (d) belong to different classes for the \textbf{Protein Subcellular dataset}. The Gaussian kernel distance for (a) and (b) is almost 0 while
for the W-PSSKM model is \textbf{4.4} (larger distance is better). The Gaussian kernel for (c) and (d) is \textbf{0.57} while for the W-PSSKM model is \textbf{0.49}  (smaller distance is better).}
 \label{fig_barplot_5959}
\end{figure}

\subsection{Inter-Class Embedding Interaction}\label{sec_inter_class_appendix}
We utilize heat maps to analyze further whether our proposed kernel can better identify different classes. These maps are generated by first taking the average of the similarity values to compute a single value for each pair of classes and then computing the pairwise cosine similarity of different class's embeddings with one another. The heat map is further normalized between [0-1] to the identity pattern. The heatmaps for the baseline Spike2Vec and its comparison with the W-PSSKM embeddings are reported in Figure~\ref{fig_heat_map_5558}, ~\ref{fig_heat_map_5959} and ~\ref{fig_heat_map_7000}. We can observe that in the case of the Spike2Vec heatmap, the embeddings for the label are similar, for all classes. This eventually means it is difficult to distinguish between different classes (as seen from Spike2Vec results in Table~\ref{tbl_results_host_classification} and~\ref{tbl_results_classification_7000} ``in the appendix") due to high pairwise similarities among their vectors. 
On the other hand, we can observe that the pairwise similarity between different class embeddings is distinguishable for W-PSSKM embeddings. This essentially means that the embeddings that belong to similar classes are highly similar to each other. In contrast, the embeddings for different classes are very different, indicating that W-PSSKM can accurately identify similar classes and different classes. This can also be verified by observing the higher predictive accuracy for the proposed method-based embeddings in Table~\ref{tbl_results_host_classification} and~\ref{tbl_results_classification_7000}.

\begin{figure}[h!]
\centering
\begin{subfigure}{.25\textwidth}
  \centering
  \includegraphics[scale = 0.120] {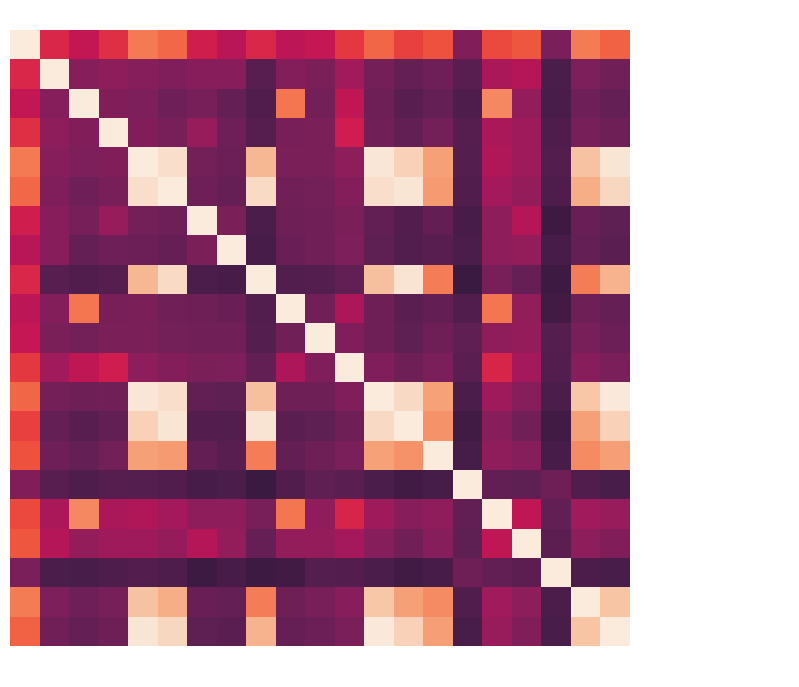}
  \caption{Spike2Vec}
\end{subfigure}%
\begin{subfigure}{.25\textwidth}
  \centering
  \includegraphics[scale = 0.120] {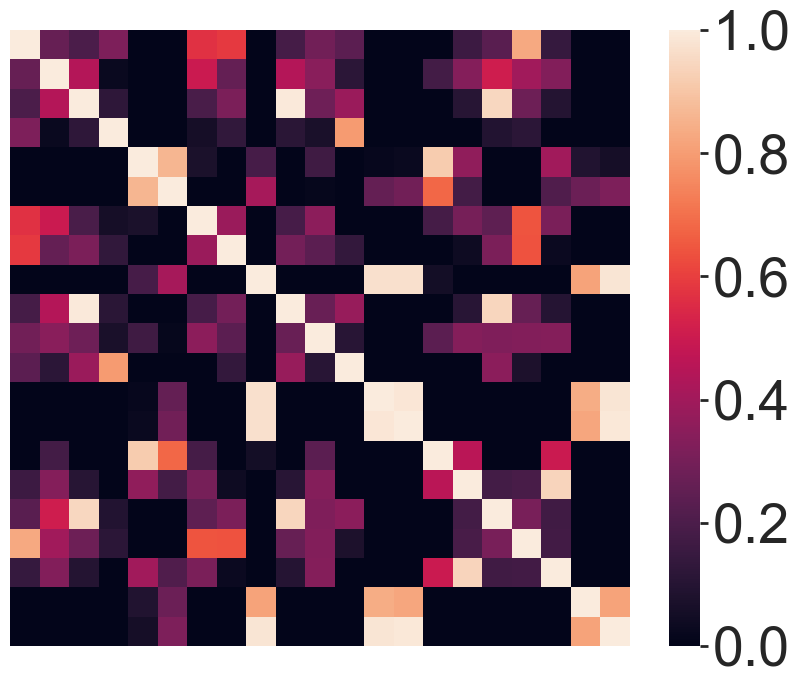}
  \caption{W-PSSKM}
\end{subfigure}
\caption{Heatmap for classes in \textbf{Coronavirus Host}. 
}
\label{fig_heat_map_5558}
\end{figure}

\begin{figure}[h!]
\centering
\begin{subfigure}{.24\textwidth}
  \centering
  \includegraphics[scale = 0.120] {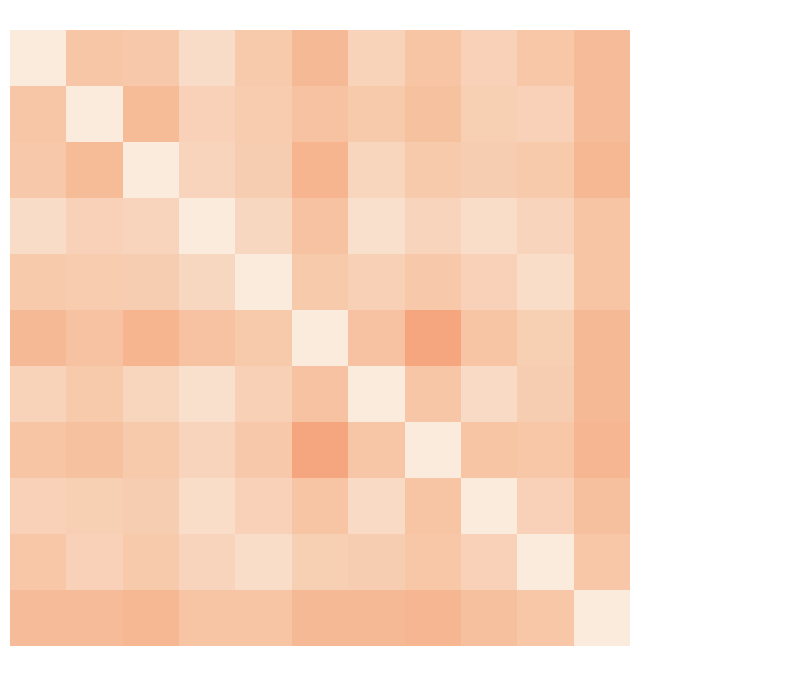}
  \caption{Spike2Vec}
\end{subfigure}%
\begin{subfigure}{.24\textwidth}
  \centering
  \includegraphics[scale = 0.120] {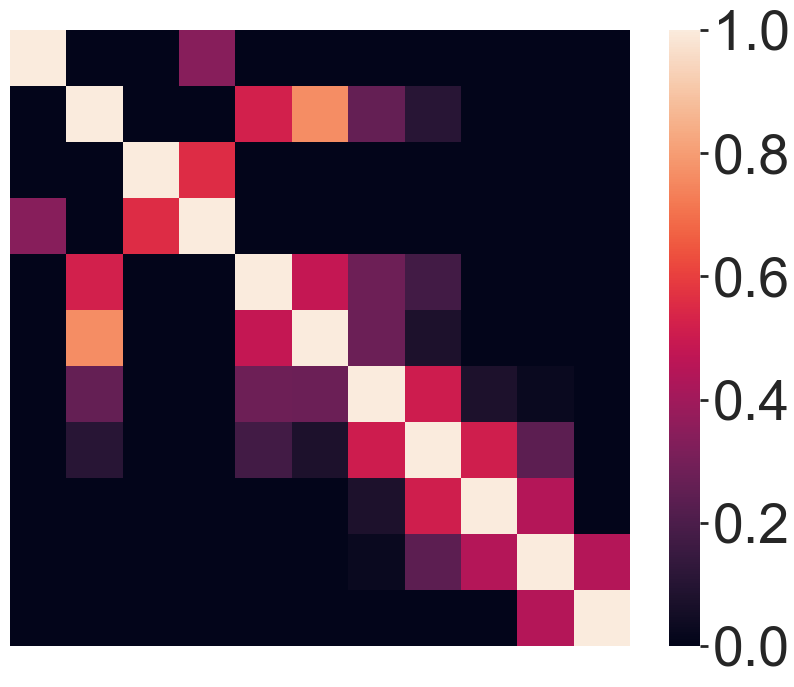}
  \caption{W-PSSKM}
\end{subfigure}
\caption{Heatmap comparison for classes in \textbf{Protein Subcellular dataset}. 
}
\label{fig_heat_map_5959}
\end{figure}

\begin{figure}[h!]
\centering
\begin{subfigure}{.24\textwidth}
  \centering
  \includegraphics[scale = 0.120] {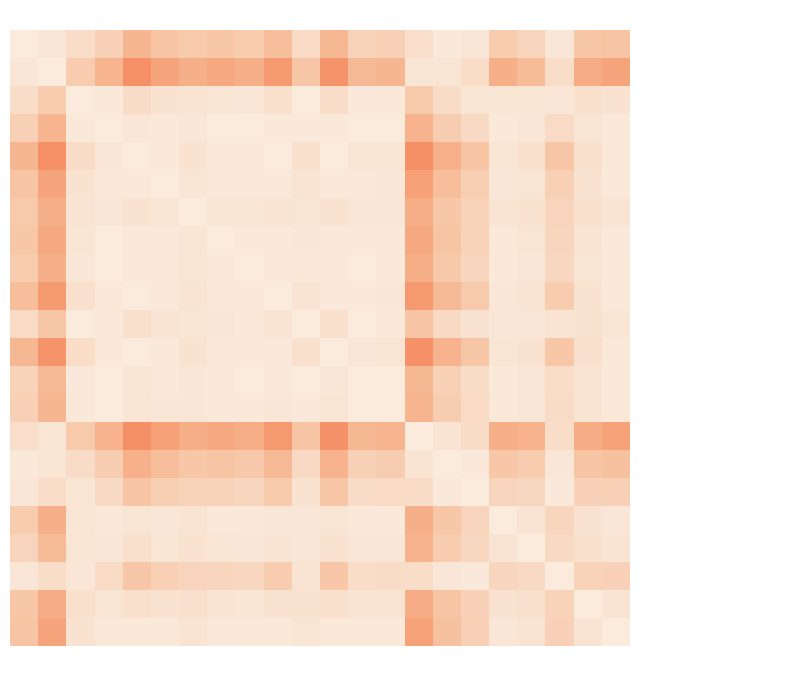}
  \caption{Spike2Vec}
\end{subfigure}%
\begin{subfigure}{.24\textwidth}
  \centering
  \includegraphics[scale = 0.120] {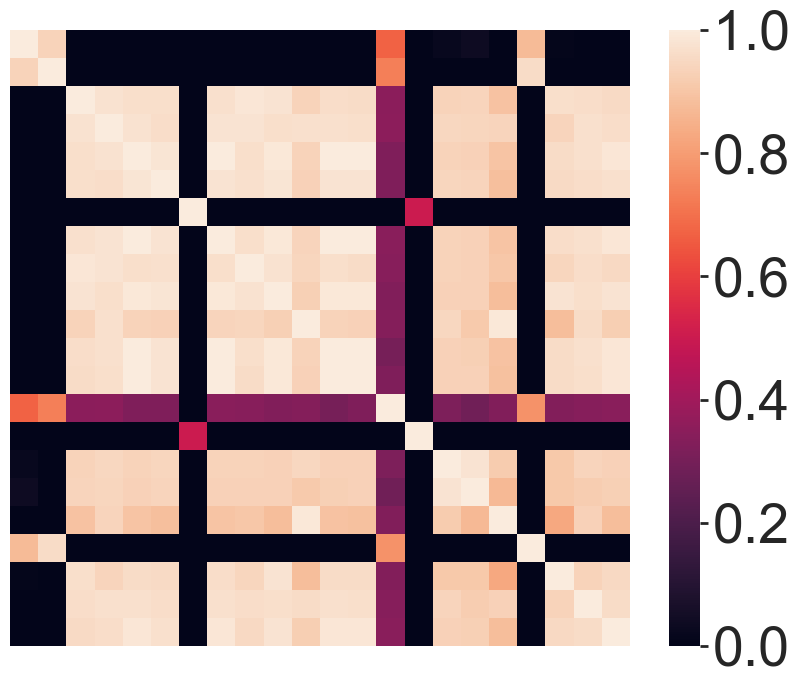}
  \caption{W-PSSKM}
\end{subfigure}
\caption{Heatmap for classes in \textbf{Spike7K dataset}. 
}
\label{fig_heat_map_7000}
\end{figure}

The discussion of the statistical significance of the results is provided in Section~\ref{sec_stats} (in appendix).

\section{Conclusion}\label{sec_conclusion}
In this paper, we propose a weighted position weight matrix-based kernel function to design a kernel matrix that we can use to perform supervised analysis of protein sequences. Using kernel PCA, we explored the classification performance of non-kernel classifiers along with the kernel SVM and showed that we could achieve higher predictive performance using classifiers such as KNN and random forest. We theoretically prove our kernel function satisfies Mercer's theorem properties, i.e., it is a continuous function with respect to $x$ and $y$, symmetric and positive semidefinite.
In the future, we will evaluate the performance of a W-PSSKM kernel for nucleotide sequences. 
Applying this method to other domains (e.g., music or video) would also be an interesting future direction.

\section{Limitations}
Despite the significant improvements and promising results achieved by the W-PSSKM in protein sequence classification, there are several limitations to consider. For example, storing the $n \times n$ dimensional kernel matrix in memory could be an issue if the value of $n$ is high. This overhead may limit the scalability of the method when applied to very large databases.
Although detailed results are shown for different protein sequence datasets, it is not clear if this method could generalize well on other biological datasets, such as nucleotide sequences and SMILES strings.

\bibliography{references}

\clearpage

\appendix

\section*{Appendix}

\section{Related Work Extended}
\label{sec_related_work_extend}
Another popular approach includes methods employing neural networks to generate numerical representations like WDGRL~\cite{shen2018wasserstein}, AutoEncoder~\cite{xie2016unsupervised}, and Evolutionary Scale Modeling or ESM-2~\cite{lin2022language} etc. WDGRL is an unsupervised technique that uses a neural network to extract numerical embeddings from the sequences. AutoEncoder follows the encoder-decoder architecture and the encoder network yields the feature embeddings for any given sequence. However, they require training data. Although ESM-2 uses advanced language models to predict protein functions, it may not emphasize evolutionary conservation to the same extent. 
A set of pre-trained models to deal with protein classification are also introduced like Protein Bert~\cite{10.1093/bioinformatics/btac020}, Seqvec~\cite{heinzinger2019modeling}, UDSMProt~\cite{strodthoff2020udsmprot}, TAPE~\cite{rao2019evaluating} etc. In Protein Bert an NN model is trained using protein sequences and this pre-trained model can be employed to get embeddings for new sequences. Likewise, SeqVec provides a pre-trained deep language model for generating protein sequence embeddings. 
In UDSMProt a universal deep sequence model is put forward which is pre-trained on unlabeled protein sequences from Swiss-Prot and further fine-tuned for the classification of proteins. The Tasks Assessing Protein Embeddings (TAPE)~\cite{rao2019evaluating} framework introduces five semi-supervised learning tasks relevant to protein biology, each with specific training, validation, and test splits to ensure meaningful biological generalization. 
However, all these methods have heavy computational costs. 
Several kernel-based analysis techniques are proposed, like gapped $k$-mer (Gkm) string kernel~\cite{ghandi2014enhanced} enables the usage of string inputs (biological sequences) for training SVMs. It determines the similarity between pairs of sequences using gapped $k$-mers, which eradicates the sparsity challenge associated with $k$-mers. 
However, the interpretation of gkmSVMs can be challenging. GkmExplain~\cite{shrikumar2019gkmexplain} is an extension of Gkm which claims to be more efficient in performance. 
The string kernel~\cite{lodhi2002text} is a kernel function that is based on the alignment of substrings in sequences but it's space inefficient. 
Likewise, in recent years, there has been growing interest in kernel functions that are based on the position-specific scoring matrix (PSSM) representation of protein sequences. The PSSM is a widely-used representation of protein sequences that encodes information about the frequency and conservation of each amino acid in a sequence relative to a set of multiple sequence alignments~\cite{henikoff1992amino}. The PSSM-based kernel functions have several advantages over other kernel functions, including their ability to capture more complex relationships between sequences. 

\begin{figure*}[h!]
  \centering
  \includegraphics[scale=0.32]{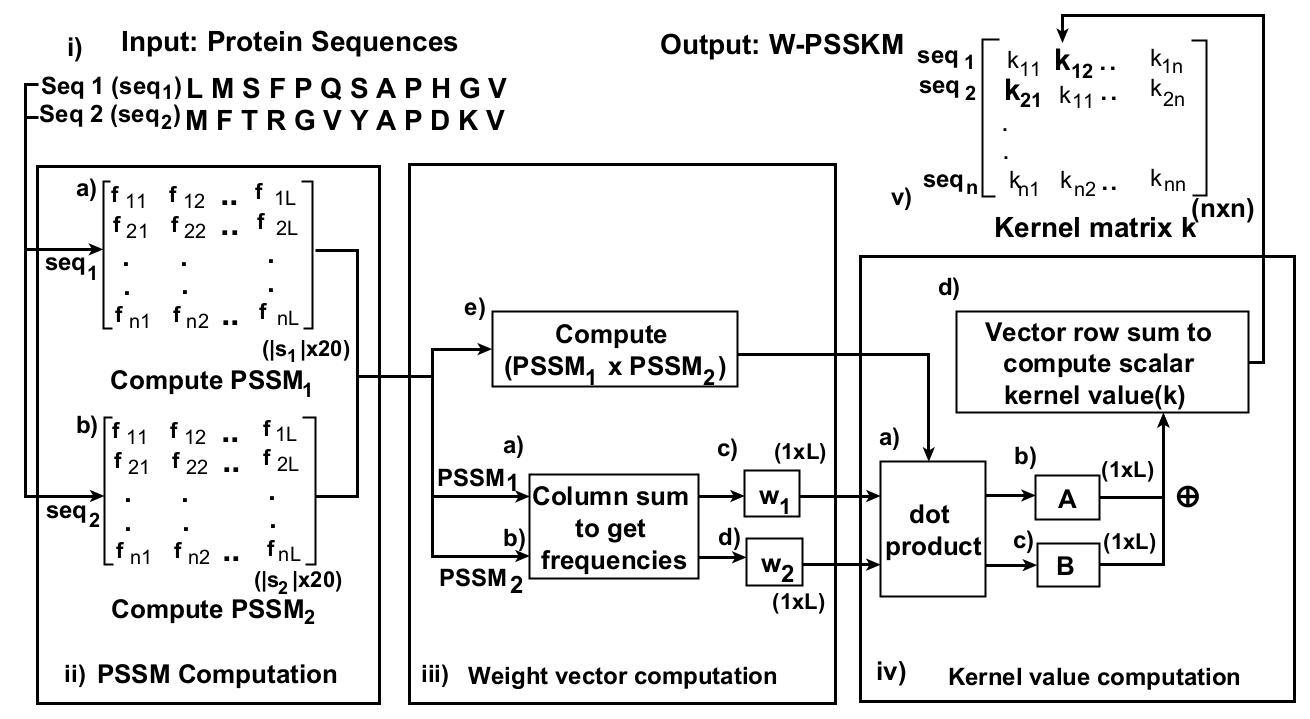}
  \caption{Flow chart for the process of constructing a W-PSSKM. 
  }
  \label{Weighted_PWM_flow_chart}
\end{figure*}

 \endgroup

\section{Runtime Analysis}
\label{sec_runtime}
Since our kernel matrix is symmetric, for each pair of sequences (i, j), where i $\leq$ j, the PSSMs are computed using Algorithm~\ref{algo_PWM_fun}, which takes $O(s \times L)$ time.
The weights are computed for each PSSM using the row-wise sum and division operation, which takes $O(s \times L)$ time.
The elementwise product of the two PSSMs is computed, which takes $O(s \times L)$ time.
The weighted dot product is then added to the kernel matrix in constant time.
Therefore, the total time complexity of computing the kernel matrix is $O(N^2 \times s \times L)$, where $N$ is the number of total sequences.
However, since the kernel matrix is symmetric, we only need to compute the upper triangle of the matrix, which reduces the time complexity by a factor of 2. Therefore, the actual time complexity of computing the kernel matrix is $O(\frac{N^2}{2}\times s \times L)$.

\section{Evaluation Metrics and ML Classifiers}\label{sec_eval_metric}
Classification is performed using Support Vector Machine (SVM), Naive Bayes (NB), Multi-Layer Perceptron (MLP), K-Nearest Neighbors (KNN), Random Forest (RF), Logistic Regression (LR), and Decision Tree (DT) ML models. The evaluation metrics used for assessing the performance of these ML models are average accuracy, precision, recall, F1 (weighted), F1 (macro), Receiver Operator Characteristic Curve Area Under the Curve (ROC AUC), and training runtime. Since we have multi-class classifications, the one-vs-rest approach is used for ROC AUC computation.

\begin{table}[h!]
    \centering
    \resizebox{0.49\textwidth}{!}{
    \begin{tabular}{p{1.4cm}cccccp{3.5cm}p{6cm}}
    \toprule
    \multirow{2}{1.2cm}{Name} & \multirow{2}{*}{$\vert$ Seq. $\vert$} & \multirow{2}{*}{$\vert$ Classes $\vert$} & \multicolumn{3}{c}{Sequence Statistics} & \multirow{2}{*}{Reference} & \multirow{2}{*}{Description} \\
    \cmidrule{4-6}
        & & & Max & Min & Mean &  &  \\
    \midrule
       \multirow{3}{1.2cm}{Spike7k}  & \multirow{3}{*}{7000} & \multirow{3}{*}{22} & \multirow{3}{*}{1274} & \multirow{3}{*}{1274} & \multirow{3}{*}{1274.00} & \multirow{3}{4cm}{~\cite{gisaid_website_url}} & The spike sequences of the SARS-CoV-2 virus having the information about the Lineage of each sequence. \\
       \midrule
       \multirow{2}{1.2cm}{Protein Subcellular}  & \multirow{2}{*}{5959} & \multirow{2}{*}{11} & \multirow{2}{*}{3678} & \multirow{2}{*}{9} & \multirow{2}{*}{326.27} &  \multirow{2}{4cm}{~\cite{ProtLoc_website_url}} & The protein sequences having information about subcellular locations. \\
       \midrule
       \multirow{4}{1.2cm}{Coronavirus Host}  & \multirow{4}{*}{5558} & \multirow{4}{*}{21} & \multirow{4}{*}{1584} & \multirow{4}{*}{9} & \multirow{4}{*}{1272.36} & \multirow{4}{3.5cm}{ViPR~\cite{pickett2012vipr}, GISAID~\cite{gisaid_website_url}} &   The spike sequences belonging to various clades of the Coronaviridae family accompanied by the infected host label e.g. Humans, Bats, Chickens, etc. \\
     \bottomrule
    \end{tabular}
    }
    \caption{Dataset Statistics for all three datasets that are used in performing the evaluation. }
    \label{tbl_data_statistics}
\end{table}

\section{Baseline and SOTA Methods}\label{apen_baselines}
This section talks about the baseline and SOTA methods used for doing evaluation in detail. 
The baseline methods include representation learning methods using feature engineering approaches such as Spike2Vec~\cite{ali2021spike2vec}, PWM2Vec~\cite{ali2022pwm2vec}, and Spaced $k$-mer~\cite{singh2017gakco}. The SOTA methods include (i) Neural Network (NN)-based models (WDGRL~\cite{shen2018wasserstein}, AutoEncoder~\cite{xie2016unsupervised}), (ii) String kernel~\cite{ali2022efficient}, (iv) End-to-End Deep Learning models (LSTM~\cite{hochreiter1997long}, CNN~\cite{lee2017human}, GRU~\cite{cho2014properties}), and (iv) Pretrained language models for protein sequences (SeqVec~\cite{heinzinger2019modeling}, Protein Bert~\cite{10.1093/bioinformatics/btac020}). The detailed description of each of these methods is given below.

\subsection{Spike2Vec}
This method uses $k$-mers concept to generate numerical embeddings of spike sequences~\cite{ali2021spike2vec}. $k$-mers are consecutive substrings of length k driven from a spike sequence. $k$-mers tend to possess the property of retaining the ordering information of a sequence, which is why they are used to get the embeddings. For a given spike sequence, the Spike2Vec method works by generating its $k$-mers first. Then using the $k$-mers it computes the frequency vector which consists of the counts of every $k$-mer. For a sequence with $\Sigma$ number of amino acids, a frequency vector of length $|\Sigma|^k$ is created where $k$ is the length of $k$-mers. The count of each $k$-mers is updated in the frequency vector by finding the respective bin of that $k$-mer in the vector. This bin-searching is a very expensive operation. The frequency vector is used as a numerical representation of the corresponding sequence. We used $k=3$ to perform our experiments.

\subsection{PWM2Vec}
This approach creates the numerical form of the biological sequences by assigning weights to the amino acids in a $k$-mer based on the location of the amino acids in the $k$-mer's position weight matrix (PWM)~\cite{ali2022pwm2vec}. It enables preserving the ordering information as well as the relative importance of amino acids. We used $k=9$ to do the experiments. 

\subsection{Spaced $k$-mer}
The high size and sparsity of feature vectors generated using $k$-mers can result in low classification performance. Therefore, a concept of spaced $k$-mer is put forward~\cite{singh2017gakco}. Spaced $k$-mers ($g$-mers) referred to non-contiguous substrings of length $k$. The $g$-mers produce low-size and less sparse feature vectors. Given any sequence, its $g$-mers are extracted, and then the $k$-mers from these $g$-mers are computed ($k<g$). However, spaced $k$-mer face the expensive bin searching challenge. For our experiments, we use $k=3$ and $g=9$.

\subsection{Wasserstein Distance Guided Representation Learning (WDGRL)}
This unsupervised technique uses a neural network to extract numerical embeddings from the sequences~\cite{shen2018wasserstein}. This domain adaption method optimizes the network by minimizing the Wasserstein distance (WD) between the given source and target networks. The OHE~\cite{kuzmin2020machine} vectors gained from the original sequences are given as input to the WDGRL, and it yields the corresponding embeddings. However, WDGRL required training data to optimize the network, which is an expensive requirement. 

\subsection{Auto-Encoder}
Autoencoder method~\cite{xie2016unsupervised} extracts the feature embeddings by employing a neural network. It works by mapping the given data from X space to a low dimensional Z space and performs the optimization iteratively. The mapping is non-linear. We define a two-layered neural network, with ADAM optimizer and MSE loss function for training it, in our experiments. This approach uses original sequences are input.

\subsection{String Kernel}
String kernel~\cite{ali2022efficient} uses a number of matched and mismatched $k$-mers to calculate the similarity between two sequences. For two $k$-mers to be considered matched, they should be at a distance $m$. It extracts the $k$-mers at $m$ distance using locality-sensitive hashing theory which is shown to reduce the computation cost. String kernel produces a square kernel matrix with dimensions equal to the count of the sequences. This kernel matrix is further given to PCA to generate reduced-dimension embeddings by us. As this method requires storing a kernel matrix, which can be space inefficient as the dataset grows larger. For experiments, we use $k=3$.

\subsection{SeqVec}
Using the ELMO (Embeddings from Language Models)~\cite{sarzynska2021detecting} language model, a vector representation of protein sequences is created in SeqVec~\cite{heinzinger2019modeling}. The generated embeddings are context-based.

\subsection{Protein Bert}
Protein Bert~\cite{10.1093/bioinformatics/btac020} is a pre-trained model introduced specifically for proteins. It's a deep language model. To handle long protein sequences efficiently and flexibly, new architectural elements are introduced in this model. The pre-training of protein Bert used around 106M protein sequences extracted from UniRef90. 

\begin{table}[h!]
    \centering
    \resizebox{0.5\textwidth}{!}{
    \begin{tabular}{p{2.5cm}p{2cm}p{10cm}}
    \toprule
   \multirow{1}{*}{Category} & \multirow{1}{*}{Method} & \multirow{1}{*}{Description} \\
    \midrule
       \multirow{8}{2.5cm}{Feature Engineering based embedding} & \multirow{2}{2cm}{Spike2Vec} & It creates the numerical feature embeddings of a sequence by taking the frequencies of its $k$-mers into account \\
       \cmidrule{2-3}
       & \multirow{4}{2cm}{PWM2Vec} & This approach creates the numerical form of the biological sequences by assigning weights to the amino acids in a $k$-mer based on the location of the amino acids in the $k$-mer's position weight matrix (PWM)\\
       \cmidrule{2-3}
       & \multirow{1}{2cm}{Spaced $k$-mer} & For a sequence, it creates the feature vector by counting the number of its spaced $k$-mers. \\
        \midrule
       \multirow{8}{2.5cm}{Neural Network based embedding} & 
       \multirow{2}{1.5cm}{WDGRL} & This unsupervised technique uses a neural network to extract numerical embeddings from the sequences. \\
       \cmidrule{2-3}
       & \multirow{3}{1.5cm}{Autoencoder} & It follows an auto-encoder architecture-based neural network to extract the embeddings. The output of the encoder contains the numerical embeddings for a given spike sequence.\\
       \cmidrule{2-3}
       & \multirow{3}{1.5cm}{ESM-2} & ESM-2 is a model trained to predict masked amino acids in protein sequences, learning from billions of such predictions to capture evolutionary information. \\
       \midrule
       \multirow{2}{2.5cm}{Kernel Function} & 
       \multirow{2}{2cm}{String Kernel} & String kernel uses a number of matched and mismatched $k$-mers to calculate the similarity between two sequences.\\
       \midrule
       \multirow{9}{2.5cm}{End-To-End Deep Learning} & 
        \multirow{3}{1.5cm}{LSTM} & 2 LSTM layer with 200 Units, Leaky Layer with alpha=0.05, dropout with 0.2 with Sigmoid Activation Function and ADAM Optimizer \\
        \cmidrule{2-3}
        & \multirow{3}{1.5cm}{CNN} & $2$ Conv. layers with 128 Filters, Kernel size 5, a Leaky Layer with alpha=0.05, a max pooling layer of size 2, a Sigmoid Activation Function along with ADAM Optimizer \\
        \cmidrule{2-3}
        & \multirow{3}{1.5cm}{GRU} & 1 GRU layer with 200 Units, a Leaky Layer with alpha=0.05, dropout with 0.2 as well as Sigmoid Activation Function with ADAM Optimizer \\
        \midrule
        \multirow{6}{2.5cm}{Pretrained Language Models} & 
        \multirow{2}{1.5cm}{SeqVec} & A pre-trained language model (using ELMO) that takes protein sequences as input and generates vector representation as output. \\
        \cmidrule{2-3}
        & \multirow{2}{2cm}{Protein Bert} & It is a deep language pre-trained model based on the transformer specifically designed for proteins. \\
        \cmidrule{2-3}
        & \multirow{1}{2cm}{TAPE} & LLM model with a self-supervised pretraining method for molecular sequence embedding generation. \\
     \bottomrule
    \end{tabular}
    }
    \caption{The summary of all the baseline methods which are used to perform the evaluation.
    }
    \label{tbl_baselines}
\end{table}

\section{Dataset Statistics}\label{apen_dataset}
A detailed discussion of each of the datasets used in the performance evaluation is covered in this section.
 
\subsection{Spike7k}
The Spike7k dataset has $7000$ spike sequences of the SARS-CoV-2 virus extracted from GISAID~\cite{gisaid_website_url}. It contains information about the $22$ unique coronavirus Lineage. The statistical distribution of this dataset is given in Table~\ref{tbl_dataset_statistics_spike7k}.

\begin{table}[h!]
    \centering
    \begin{tabular}{lc|lc}
    \toprule
        Lineage & Seq Count & Lineage & Seq Count \\
        \midrule \midrule
        B.1.1.7 & 3369 & B.1.617.2 & 875 \\
        AY.4 & 593 & B.1.2 & 333 \\
        B.1 & 292 & B.1.177 & 243 \\
        P.1 & 194 & B.1.1 & 163 \\
        B.1.429 & 107 & B.1.526 & 104 \\
        AY.12 & 101 & B.1.160 & 92 \\
        B.1.351 & 81 & B.1.427 & 65 \\
        B.1.1.214 & 64 & B.1.1.519 & 56 \\
        D.2 & 55 & B.1.221 & 52 \\
        B.1.177.21 & 47 & B.1.258 & 46 \\
        B.1.243 & 36 & R.1 & 32 \\
        \midrule
       - & - & Total & 7000\\
        \bottomrule
    \end{tabular}
    \caption{Dataset Statistics for Spike7k data. It shows the count of sequences for each Lineage in the dataset.}
    \label{tbl_dataset_statistics_spike7k}
\end{table}

\subsection{Coronavirus Host}
This dataset refers to a set of spike sequences belonging to various clades of the Coronaviridae family accompanied by the infected host information for each spike sequence. It has $5558$ sequences containing the information about $21$ unique hosts, which are extracted from GISAID~\cite{gisaid_website_url} and ViPR~\cite{pickett2012vipr} databased~\cite{ali2022pwm2vec}. The statistical distribution of host data is given in Table~\ref{tbl_dataset_statistics_host}.

\begin{table}[h!]
    \centering
    \resizebox{0.49\textwidth}{!}{
    \begin{tabular}{lc|lc}
    \toprule
        Host & Seq Count & Host & Seq Count \\
        \midrule \midrule
        Bats & 153 & Bovines & 88 \\
        Cats & 123 & Cattle & 1 \\
        Equine & 5 & Fish & 2 \\
        Humans & 1813 & Pangolins & 21 \\
        Rats & 26 & Turtle & 1 \\
        Weasel & 994 & Birds & 374 \\
        Camels & 297 & Canis & 40 \\
        Dolphins & 7 & Environment & 1034 \\
        Hedgehog & 15 & Monkey & 2 \\
        Python & 2 & Swines & 558 \\
        Unknown & 2 & & \\
        \midrule
      - & - & Total & 5558\\
        \bottomrule
    \end{tabular}
    }
    \caption{Dataset Statistics for Host data. It shows the count of sequences for each host in the dataset.}
    \label{tbl_dataset_statistics_host}
\end{table}

\subsection{Protein Subcellular}
Protein Subcellular data~\cite{ProtLoc_website_url} consists of unaligned protein sequences having information about $11$ unique subcellular locations, which are used as the class labels for the classification task. It has $5959$
sequences. The subcellular locations with their respective counts are shown in Table~\ref{tbl_dataset_statistics_protLoc}.

\begin{table}[h!]
    \centering
    \begin{tabular}{lc}
    \toprule
        Subcellular Locations & Num. of Sequences \\
        \midrule \midrule
        Cytoplasm &  1411 \\ 
        Plasma Membrane & 1238 \\
        Extracellular Space & 843 \\
        Nucleus & 837 \\ 
        Mitochondrion & 510 \\
        Chloroplast & 449  \\ 
        Endoplasmic Reticulum & 198 \\ 
        Peroxisome & 157 \\ 
        Golgi Apparatus & 150 \\ 
        Lysosomal & 103 \\ 
        Vacuole & 63 \\ 
        \midrule
        Total & 5959 \\
        \bottomrule
    \end{tabular}
    \caption{Dataset Statistics for Protein Subcellular data. It shows the count of sequences for each subcellular location in the dataset.}
    \label{tbl_dataset_statistics_protLoc}
\end{table}

\section{Results}

\subsection{Confusion Matrix}
The confusion matrix for the ESM-2-based model and our proposed method on the Protein Subcellular dataset using the Random Forest classifier is shown in Table~\ref{Table_Confusion_Matrix_Prot_Sub_ESM-2} and Table~\ref{Table_Confusion_Matrix_Prot_Sub_W_PWM}.
We can observe from the above confusion matrices that the proposed kernel method generally shows improved accuracy in most categories, notably in chloroplast and peroxisomal classifications. This could be because when our proposed method focuses on conserved sequences, it is better at identifying functional domains crucial for the localization of proteins to specific organelles. For example, certain signal peptides or transit peptides are critical for targeting proteins in the chloroplast, mitochondria, or nucleus. The ESM-2 may not differentiate these functional motifs precisely, leading to higher misclassification rates. The proteins targeted to specific organelles often have highly conserved regions that are critical for their function and localization. PSSMs are designed to detect these conserved regions, making them more sensitive to evolutionary signals that guide protein localization. Although ESM-2 uses advanced language models to predict protein functions, it may not emphasize evolutionary conservation to the same extent.

\begin{table*}[h!]
  \centering
    \resizebox{0.99\textwidth}{!}{
\begin{tabular}{|c|c|c|c|c|c|c|c|c|c|c|c|}
\hline & ER & Golgi & chloroplast & cytoplasmic & extracellular & lysosomal & mitochondrial & nuclear & peroxisomal & plasma membrane & vacuolar \\
\hline ER & 63 & 0 & 0 & 0 & 0 & 0 & 0 & 0 & 0 & 0 & 0 \\
\hline Golgi & 0 & 43 & 0 & 0 & 0 & 0 & 0 & 0 & 0 & 0 & 0 \\
\hline chloroplast & 0 & 0 & 80 & 56 & 0 & 0 & 0 & 0 & 0 & 0 & 0 \\
\hline cytoplasmic & 0 & 0 & 15 & 410 & 0 & 0 & 0 & 0 & 0 & 0 & 0 \\
\hline extracellular & 0 & 0 & 0 & 0 & 227 & 0 & 26 & 0 & 0 & 0 & 0 \\
\hline lysosomal & 0 & 0 & 0 & 0 & 0 & 31 & 0 & 0 & 0 & 0 & 0 \\
\hline mitochondrial & 0 & 0 & 0 & 0 & 28 & 0 & 134 & 0 & 0 & 0 & 0 \\
\hline nuclear & 0 & 0 & 0 & 0 & 0 & 0 & 0 & 248 & 0 & 0 & 0 \\
\hline peroxisomal & 0 & 0 & 2 & 38 & 0 & 0 & 0 & 0 & 1 & 0 & 0 \\
\hline plasma membrane & 0 & 0 & 0 & 0 & 0 & 0 & 0 & 0 & 0 & 367 & 0 \\
\hline vacuolar & 0 & 0 & 0 & 0 & 0 & 0 & 0 & 0 & 0 & 0 & 19 \\
\hline
\end{tabular}
 }
  \caption{Confusion Matrix for ESM-2-based model for the \textbf{Protein Subcellular dataset}.}
  \label{Table_Confusion_Matrix_Prot_Sub_ESM-2}
\end{table*}

\begin{table*}[h!]
  \centering
    \resizebox{0.99\textwidth}{!}{
\begin{tabular}{|c|c|c|c|c|c|c|c|c|c|c|c|}
\hline & ER & Golgi & chloroplast & cytoplasmic & extracellular & lysosomal & mitochondrial & nuclear & peroxisomal & plasma membrane & vacuolar \\
\hline ER & 59 & 0 & 0 & 4 & 0 & 0 & 0 & 0 & 0 & 0 & 0 \\
\hline Golgi & 0 & 45 & 0 & 0 & 0 & 1 & 0 & 0 & 0 & 0 & 0 \\
\hline chloroplast & 0 & 0 & 138 & 1 & 0 & 0 & 0 & 0 & 0 & 0 & 0 \\
\hline cytoplasmic & 0 & 0 & 0 & 452 & 0 & 0 & 0 & 0 & 0 & 0 & 0 \\
\hline extracellular & 0 & 3 & 0 & 5 & 256 & 0 & 0 & 0 & 0 & 0 & 0 \\
\hline lysosomal & 0 & 0 & 0 & 1 & 1 & 24 & 2 & 0 & 0 & 0 & 0 \\
\hline mitochondrial & 0 & 0 & 0 & 0 & 0 & 0 & 146 & 0 & 0 & 0 & 0 \\
\hline nuclear & 0 & 0 & 0 & 0 & 0 & 0 & 2 & 247 & 0 & 0 & 0 \\
\hline peroxisomal & 0 & 0 & 0 & 0 & 0 & 0 & 0 & 1 & 36 & 1 & 0 \\
\hline plasma membrane & 0 & 0 & 0 & 0 & 0 & 0 & 0 & 0 & 1 & 341 & 0 \\
\hline vacuolar & 0 & 0 & 0 & 0 & 0 & 0 & 0 & 0 & 0 & 1 & 20 \\
\hline
\end{tabular}
 }
  \caption{Confusion Matrix for our proposed method for the \textbf{Protein Subcellular dataset}.}
  \label{Table_Confusion_Matrix_Prot_Sub_W_PWM}
\end{table*}

\subsection{Statistical Significance}\label{sec_stats}
To evaluate the stability and relevance of classification results, we computed p-values (for all methods) using the average and standard deviations (std), see Tables~\ref{tbl_std_org_host} and ~\ref{tbl_results_classification_std}. of all evaluation metrics (for $5$ runs of experiments) for all three datasets. The p-values for all pairwise comparisons of the proposed model with baselines (and SOTA) were $<0.05$ for all but one evaluation metric (training runtime), showing the statistical significance of the results. For training runtime, since we can see from std. results (in appendix) that there is a comparatively higher variation in the values, the p-values were sometimes $>0.05$. This is due to the fact the training time can be affected by many factors, such as processor performance and the number of jobs at any given time.

The standard deviation results for Coronavirus Host data are reported in Table~\ref{tbl_std_org_host}. We can observe that overall the std. values are very small for all evaluation metrics.

\begin{table}[h!]
  \centering
    \resizebox{0.49\textwidth}{!}{
  \begin{tabular}{p{2.6cm}p{1.3cm}p{1.3cm}p{1.3cm}p{1.3cm}p{1.3cm}cp{1.3cm} | p{1.3cm}}
    \toprule
    \multirow{3}{1.1cm}{Embed. Method} & \multirow{3}{0.7cm}{ML Algo.} & \multirow{3}{*}{Acc.} & \multirow{3}{*}{Prec.} & \multirow{3}{*}{Recall} & \multirow{3}{0.9cm}{F1 weigh.} & \multirow{3}{0.9cm}{F1 Macro} & \multirow{3}{1.2cm}{ROC- AUC} & Train. runtime (sec.) \\	
    \midrule \midrule	
    
    \multirow{7}{*}{Spike2Vec}  
     & SVM & 0.02187 & 0.03118 & 0.02187 & 0.02506 & 0.01717 & 0.01059 & 0.52562 \\
 & NB & 0.00954 & 0.03743 & 0.00954 & 0.01682 & 0.01232 & 0.00833 & 0.00818 \\
 & MLP & 0.00295 & 0.00383 & 0.00295 & 0.00393 & 0.00025 & 0.00284 & 6.35600 \\
 & KNN & 0.01892 & 0.02670 & 0.01892 & 0.02128 & 0.02197 & 0.01347 & 0.01097 \\
 & RF & 0.00898 & 0.00720 & 0.00898 & 0.00712 & 0.00227 & 0.00176 & 0.04175 \\
 & LR & 0.00802 & 0.01287 & 0.00802 & 0.01179 & 0.00698 & 0.00363 & 0.03516 \\
 & DT & 0.01608 & 0.01437 & 0.01608 & 0.01533 & 0.01894 & 0.00981 & 0.00880 \\
    \cmidrule{2-9}	
    \multirow{7}{*}{PWM2Vec}  
     & SVM & 0.01459 & 0.01735 & 0.01459 & 0.01737 & 0.01599 & 0.01051 & 0.42070 \\
 & NB & 0.01808 & 0.02434 & 0.01808 & 0.02163 & 0.01882 & 0.00905 & 0.00863 \\
 & MLP & 0.01898 & 0.02002 & 0.01898 & 0.01990 & 0.01293 & 0.00793 & 7.39342 \\
 & KNN & 0.01098 & 0.01342 & 0.01098 & 0.01097 & 0.00904 & 0.00441 & 0.04144 \\
 & RF & 0.02330 & 0.01223 & 0.02330 & 0.02348 & 0.01934 & 0.01331 & 0.04276 \\
 & LR & 0.01896 & 0.01959 & 0.01896 & 0.02155 & 0.01807 & 0.01092 & 0.00840 \\
 & DT & 0.00974 & 0.01461 & 0.00974 & 0.01171 & 0.00781 & 0.00558 & 0.01814 \\
    \cmidrule{2-9}	
    \multirow{7}{*}{String Kernel}  
 & SVM & 0.00892 & 0.00545 & 0.00892 & 0.00737 & 0.00150 & 0.01176 & 0.08293 \\
 & NB & 0.03446 & 0.04765 & 0.03446 & 0.03270 & 0.02739 & 0.01621 & 0.00054 \\
 & MLP & 0.02197 & 0.06306 & 0.02197 & 0.02880 & 0.02930 & 0.01528 & 3.28975 \\
 & KNN & 0.01546 & 0.01811 & 0.01546 & 0.01752 & 0.01364 & 0.00600 & 0.00257 \\
 & RF & 0.02143 & 0.03719 & 0.02143 & 0.02293 & 0.02613 & 0.01234 & 0.07206 \\
 & LR & 0.00898 & 0.03013 & 0.00898 & 0.01760 & 0.01914 & 0.00551 & 0.00171 \\
 & DT & 0.02911 & 0.03146 & 0.02911 & 0.03035 & 0.03428 & 0.01948 & 0.00829 \\
     \cmidrule{2-9}	
    \multirow{7}{*}{WDGRL}  
    & SVM & 0.008378 & 0.005078 & 0.008378 & 0.006888 & 0.00141 & 0.002417 & 0.034498 \\
 & NB & 0.008720 & 0.055455 & 0.00872 & 0.022475 & 0.022506 & 0.007747 & 0.000352 \\
 & MLP & 0.016103 & 0.010655 & 0.016103 & 0.018511 & 0.014246 & 0.007622 & 2.437331 \\
 & KNN & 0.010047 & 0.009635 & 0.010047 & 0.010275 & 0.011106 & 0.006126 & 0.002751 \\
 & RF & 0.013395 & 0.019497 & 0.013395 & 0.015266 & 0.018384 & 0.008316 & 0.035652 \\
 & LR & 0.007675 & 0.094971 & 0.007675 & 0.008903 & 0.005274 & 0.00175 & 0.001188 \\
 & DT & 0.009280 & 0.008941 & 0.00928 & 0.009266 & 0.007579 & 0.004472 & 0.004392 \\
     \cmidrule{2-9}	
    \multirow{7}{*}{Spaced Kernel}  
     & SVM & 0.00908 & 0.00555 & 0.00908 & 0.00751 & 0.00152 & 0.01198 & 0.08447 \\
 & NB & 0.03510 & 0.04853 & 0.03510 & 0.03330 & 0.02790 & 0.01651 & 0.00055 \\
 & MLP & 0.02238 & 0.06423 & 0.02238 & 0.02934 & 0.02984 & 0.01557 & 3.35067 \\
 & KNN & 0.01575 & 0.01845 & 0.01575 & 0.01785 & 0.01389 & 0.00611 & 0.00262 \\
 & RF & 0.02183 & 0.03788 & 0.02183 & 0.02336 & 0.02662 & 0.01257 & 0.07340 \\
 & LR & 0.00915 & 0.03069 & 0.00915 & 0.01793 & 0.01949 & 0.00561 & 0.00174 \\
 & DT & 0.02965 & 0.03204 & 0.02965 & 0.03091 & 0.03492 & 0.01984 & 0.00845 \\
\cmidrule{2-9}	
    \multirow{7}{1.9cm}{Autoencoder}  
  & SVM & 0.00956 & 0.00974 & 0.00956 & 0.01059 & 0.00127 & 0.00094 & 0.35010 \\
 & NB & 0.05871 & 0.04630 & 0.05871 & 0.05414 & 0.02578 & 0.01233 & 0.03049 \\
 & MLP & 0.00846 & 0.01142 & 0.00846 & 0.00776 & 0.01468 & 0.00882 & 1.00276 \\
 & KNN & 0.00631 & 0.00803 & 0.00631 & 0.00807 & 0.00699 & 0.00493 & 0.01380 \\
 & RF & 0.00338 & 0.00548 & 0.00338 & 0.00381 & 0.01029 & 0.00786 & 0.72100 \\
 & LR & 0.00982 & 0.00982 & 0.00982 & 0.01072 & 0.00128 & 0.00093 & 0.19975 \\
 & DT & 0.01025 & 0.00968 & 0.01025 & 0.00968 & 0.02163 & 0.00821 & 0.09998 \\

\cmidrule{2-9}	
    \multirow{7}{1.9cm}{SeqVec}  
  & SVM & 0.00729 & 0.00924 & 0.00729 & 0.00924 & 0.00102 & 0.00031 & 0.29267 \\
 & NB & 0.14408 & 0.06203 & 0.14408 & 0.11723 & 0.02721 & 0.01650 & 0.01298 \\
 & MLP & 0.01185 & 0.01247 & 0.01185 & 0.01129 & 0.02103 & 0.00999 & 0.68591 \\
 & KNN & 0.01281 & 0.01485 & 0.01281 & 0.01456 & 0.02329 & 0.01131 & 0.05557 \\
 & RF & 0.01050 & 0.01599 & 0.01050 & 0.01294 & 0.01490 & 0.00771 & 0.36725 \\
 & LR & 0.00795 & 0.00984 & 0.00795 & 0.01007 & 0.00126 & 0.00053 & 0.22741 \\
 & DT & 0.01119 & 0.01183 & 0.01119 & 0.01277 & 0.02537 & 0.00989 & 0.12363 \\

     \cmidrule{2-9}	
\multirow{7}{1.9cm}{ESM-2}  
 & SVM & 0.002996 & 0.004791 & 0.002996 & 0.003504 & 0.004206 & 0.002321 & 63.325091 \\
 & NB & 0.010739 & 0.00679 & 0.010739 & 0.012877 & 0.005521 & 0.002957 & 1.637343 \\
 & MLP & 0.005551 & 0.005859 & 0.005551 & 0.004859 & 0.006128 & 0.002757 & 37.433558 \\
 & KNN & 0.004605 & 0.00998 & 0.004605 & 0.005275 & 0.015125 & 0.006012 & 1.023186 \\
 & RF & 0.00622 & 0.018651 & 0.00622 & 0.007063 & 0.006244 & 0.003455 & 2.860571 \\
 & LR & 0.005202 & 0.006391 & 0.005202 & 0.005831 & 0.003528 & 0.001892 & 24.725468 \\
 & DT & 0.011207 & 0.010541 & 0.011207 & 0.010715 & 0.007195 & 0.003521 & 3.567783 \\
 \cmidrule{2-9}
  \multirow{7}{1.6cm}{TAPE}
   & SVM & 0.005941 & 0.010001 & 0.005941 & 0.005494 & 0.060968 & 0.023419 & 0.137464 \\
 & NB & 0.009549 & 0.010933 & 0.009549 & 0.01013 & 0.044898 & 0.020246 & 0.023094 \\
 & MLP & 0.016546 & 0.010195 & 0.016546 & 0.01605 & 0.04501 & 0.026421 & 0.567482 \\
 & KNN & 0.012213 & 0.015524 & 0.012213 & 0.011534 & 0.050628 & 0.029621 & 0.009386 \\
 & RF & 0.008764 & 0.012207 & 0.008764 & 0.008546 & 0.054582 & 0.030222 & 0.465312 \\
 & LR & 0.006207 & 0.005478 & 0.006207 & 0.00697 & 0.062016 & 0.032901 & 0.530472 \\
 & DT & 0.009587 & 0.008818 & 0.009587 & 0.009863 & 0.050643 & 0.027967 & 0.358515 \\
     \cmidrule{2-9}
    \multirow{7}{1.9cm}{Weighted PSSKM Kernel}  
      & SVM & 0.00459 & 0.00418 & 0.00459 & 0.00435 & 0.04749 & 0.01768 & 0.07872 \\
 & NB & 0.02022 & 0.01953 & 0.02022 & 0.01775 & 0.05769 & 0.02999 & 0.02715 \\
 & MLP & 0.00591 & 0.00586 & 0.00591 & 0.00583 & 0.03420 & 0.02921 & 3.88783 \\
 & KNN & 0.00544 & 0.00658 & 0.00544 & 0.00567 & 0.06500 & 0.03523 & 0.02241 \\
 & RF & 0.00283 & 0.00290 & 0.00283 & 0.00329 & 0.03443 & 0.02141 & 0.18349 \\
 & LR & 0.00263 & 0.00331 & 0.00263 & 0.00323 & 0.05910 & 0.02767 & 1.81866 \\
 & DT & 0.00510 & 0.00403 & 0.00510 & 0.00444 & 0.03242 & 0.02751 & 0.05445 \\

    \bottomrule
  \end{tabular}
  }
  \caption{Standard Deviation values of 5 runs for Classification results on the proposed and SOTA methods for \textbf{Coronavirus Host dataset}.}
  \label{tbl_std_org_host}
\end{table}

The standard deviation results for Spike7k and Protein Subcellular data are given in Table~\ref{tbl_results_classification_std}. We can again observe that overall the std. values are very small for all evaluation metrics.
\begin{table*}[h!]
\centering
\resizebox{1\textwidth}{!}{
 \begin{tabular}{@{\extracolsep{6pt}}p{2.5cm}lp{1.1cm}p{1.1cm}p{1.1cm}p{1.3cm}p{1.3cm}p{1.1cm}p{1.7cm}
 p{1.1cm}p{1.1cm}p{1.1cm}p{1.3cm}p{1.3cm}p{1.1cm}p{1.7cm}}
    \toprule
    & & \multicolumn{7}{c}{Spike7k} & \multicolumn{7}{c}{Protein Subcellular} \\
    \cmidrule{3-9} \cmidrule{10-16}
        \multirow{2}{*}{Embeddings} & \multirow{2}{*}{Algo.} & \multirow{2}{*}{Acc. $\uparrow$} & \multirow{2}{*}{Prec. $\uparrow$} & \multirow{2}{*}{Recall $\uparrow$} & \multirow{2}{1.4cm}{F1 (Weig.) $\uparrow$} & \multirow{2}{1.5cm}{F1 (Macro) $\uparrow$} & \multirow{2}{1.2cm}{ROC AUC $\uparrow$} & Train Time (sec.) $\downarrow$
          & \multirow{2}{*}{Acc. $\uparrow$} & \multirow{2}{*}{Prec. $\uparrow$} & \multirow{2}{*}{Recall $\uparrow$} & \multirow{2}{1.4cm}{F1 (Weig.) $\uparrow$} & \multirow{2}{1.5cm}{F1 (Macro) $\uparrow$} & \multirow{2}{1.2cm}{ROC AUC $\uparrow$} & Train Time (sec.) $\downarrow$\\
        \midrule \midrule
        \multirow{7}{1.2cm}{Spike2Vec}
 & SVM & 0.005339 & 0.005547 & 0.005339 & 0.004908 & 0.00894 & 0.003092 & 3.916014  & 0.01981 & 0.02825 & 0.01981 & 0.02270 & 0.01555 & 0.00959 & 0.47610 \\
 & NB & 0.165977 & 0.047923 & 0.165977 & 0.124314 & 0.028412 & 0.021885 & 0.234008 & 0.00864 & 0.03391 & 0.00864 & 0.01524 & 0.01116 & 0.00755 & 0.00741 \\
 & MLP & 0.010331 & 0.011387 & 0.010331 & 0.010000 & 0.024026 & 0.013581 & 5.913928 & 0.00267 & 0.00347 & 0.00267 & 0.00356 & 0.00022 & 0.00257 & 5.75725 \\
 & KNN & 0.015266 & 0.00962 & 0.015266 & 0.014351 & 0.016323 & 0.006342 & 2.590869 & 0.01714 & 0.02418 & 0.01714 & 0.01927 & 0.01990 & 0.01220 & 0.00993 \\
 & RF & 0.007615 & 0.010725 & 0.007615 & 0.008411 & 0.016425 & 0.007389 & 0.433674 & 0.00814 & 0.00652 & 0.00814 & 0.00645 & 0.00206 & 0.00159 & 0.03782 \\
 & LR & 0.006048 & 0.005895 & 0.006048 & 0.006385 & 0.01425 & 0.005388 & 2.752432 & 0.00726 & 0.01165 & 0.00726 & 0.01068 & 0.00633 & 0.00329 & 0.03185 \\
 & DT & 0.004937 & 0.005132 & 0.004937 & 0.004463 & 0.003758 & 0.002974 & 0.437479 & 0.01457 & 0.01302 & 0.01457 & 0.01389 & 0.01716 & 0.00889 & 0.00797 \\
        \cmidrule{2-9} \cmidrule{10-16}
        \multirow{7}{1.2cm}{PWM2Vec}
& SVM & 0.009732 & 0.014177 & 0.009732 & 0.011394 & 0.040983 & 0.022625 & 0.111355 & 0.01386 & 0.01648 & 0.01386 & 0.01651 & 0.01519 & 0.00998 & 0.39967 \\
 & NB & 0.04711 & 0.028739 & 0.04711 & 0.034415 & 0.021785 & 0.019178 & 0.047905 & 0.01718 & 0.02313 & 0.01718 & 0.02055 & 0.01788 & 0.00860 & 0.00820 \\
 & MLP & 0.004077 & 0.010426 & 0.004077 & 0.005091 & 0.020858 & 0.007274 & 2.914309 & 0.01803 & 0.01902 & 0.01803 & 0.01891 & 0.01228 & 0.00753 & 7.02375 \\
 & KNN & 0.011276 & 0.010159 & 0.011276 & 0.011393 & 0.017793 & 0.007173 & 0.048912 & 0.01043 & 0.01275 & 0.01043 & 0.01042 & 0.00859 & 0.00419 & 0.03936 \\
 & RF & 0.004304 & 0.01031 & 0.004304 & 0.007477 & 0.00978 & 0.007079 & 0.784592 & 0.02213 & 0.01162 & 0.02213 & 0.02231 & 0.01838 & 0.01264 & 0.04062 \\
 & LR & 0.003573 & 0.00751 & 0.003573 & 0.004969 & 0.010045 & 0.005569 & 5.316077 & 0.01802 & 0.01861 & 0.01802 & 0.02047 & 0.01716 & 0.01037 & 0.00798 \\
 & DT & 0.005904 & 0.00566 & 0.005904 & 0.004654 & 0.012841 & 0.005028 & 0.058772 & 0.00925 & 0.01388 & 0.00925 & 0.01112 & 0.00742 & 0.00530 & 0.01723 \\
         \cmidrule{2-9} \cmidrule{10-16}
        \multirow{7}{1.9cm}{String Kernel}
 & SVM & 0.010442 & 0.013078 & 0.010442 & 0.011673 & 0.022631 & 0.007962 & 0.24096 & 0.00952 & 0.00581 & 0.00952 & 0.00786 & 0.00160 & 0.01255 & 0.08849 \\
 & NB & 0.008815 & 0.005557 & 0.008815 & 0.00703 & 0.011761 & 0.009719 & 0.006103 & 0.03677 & 0.05084 & 0.03677 & 0.03489 & 0.02923 & 0.01730 & 0.00058 \\
 & MLP & 0.007526 & 0.010713 & 0.007526 & 0.007969 & 0.020091 & 0.00948 & 2.020901 & 0.02344 & 0.06729 & 0.02344 & 0.03073 & 0.03126 & 0.01631 & 4.14838 \\
 & KNN & 0.006269 & 0.006129 & 0.006269 & 0.005595 & 0.014096 & 0.00755 & 0.050194 & 0.01650 & 0.01933 & 0.01650 & 0.01870 & 0.01455 & 0.00640 & 0.00274 \\
 & RF & 0.008558 & 0.010725 & 0.008558 & 0.011033 & 0.01859 & 0.006795 & 0.194081 & 0.02286 & 0.03968 & 0.02286 & 0.02447 & 0.02789 & 0.01317 & 0.07689 \\
 & LR & 0.006258 & 0.013144 & 0.006258 & 0.007598 & 0.009922 & 0.00391 & 0.285232 & 0.00959 & 0.03215 & 0.00959 & 0.01878 & 0.02042 & 0.00588 & 0.00183 \\
 & DT & 0.00958 & 0.011658 & 0.00958 & 0.01026 & 0.019321 & 0.007745 & 0.034749 & 0.03106 & 0.03357 & 0.03106 & 0.03238 & 0.03658 & 0.02079 & 0.00885 \\
          \cmidrule{2-9} \cmidrule{10-16}
           \multirow{7}{1.2cm}{WDGRL}  
     & SVM & 0.006843 & 0.007175 & 0.006843 & 0.007533 & 0.018218 & 0.006985 & 0.127963  & 0.005067 & 0.00313 & 0.005067 & 0.004219 & 0.000845 & 0.007464 & 0.036333 \\
 & NB & 0.012983 & 0.014996 & 0.012983 & 0.013997 & 0.009849 & 0.005343 & 0.002441 & 0.018709 & 0.028927 & 0.018709 & 0.012416 & 0.009551 & 0.007475 & 0.000338 \\
 & MLP & 0.009597 & 0.016199 & 0.009597 & 0.008501 & 0.009785 & 0.003415 & 1.973822 & 0.00703 & 0.037114 & 0.00703 & 0.01059 & 0.011773 & 0.006923 & 0.932333 \\
 & KNN & 0.012222 & 0.006539 & 0.012222 & 0.011150 & 0.012703 & 0.007168 & 0.027236 & 0.00703 & 0.005878 & 0.00703 & 0.006132 & 0.003794 & 0.002335 & 0.001434 \\
 & RF & 0.004658 & 0.002715 & 0.004658 & 0.004319 & 0.014516 & 0.011376 & 0.128241 & 0.008417 & 0.011911 & 0.008417 & 0.008514 & 0.006176 & 0.003087 & 0.044674 \\
 & LR & 0.005999 & 0.006079 & 0.005999 & 0.005070 & 0.010821 & 0.008120 & 0.062362 & 0.005067 & 0.016623 & 0.005067 & 0.011562 & 0.013997 & 0.004144 & 0.000185 \\
 & DT & 0.002869 & 0.003766 & 0.002869 & 0.001588 & 0.017623 & 0.012498 & 0.001499 & 0.017918 & 0.01796 & 0.017918 & 0.018219 & 0.016001 & 0.009088 & 0.002927 \\
 \cmidrule{2-9} \cmidrule{10-16}
 \multirow{7}{1.9cm}{Spaced $k$-mers} 
     & SVM & 0.014645 & 0.012340 & 0.014645 & 0.015661 & 0.024540 & 0.014596 & 1.106693  & 0.008258 & 0.005042 & 0.008258 & 0.006823 & 0.001385 & 0.010888 & 0.076791 \\
 & NB & 0.041749 & 0.023858 & 0.041749 & 0.033381 & 0.025358 & 0.008237 & 0.036856 & 0.031911 & 0.044116 & 0.031911 & 0.030275 & 0.025365 & 0.01501 & 0.00154 \\
 & MLP & 0.006472 & 0.008860 & 0.006472 & 0.003471 & 0.022909 & 0.010423 & 3.305528 & 0.020343 & 0.058389 & 0.020343 & 0.02667 & 0.027129 & 0.014152 & 3.046066 \\
 & KNN & 0.012241 & 0.018108 & 0.012241 & 0.013407 & 0.037266 & 0.019443 & 0.089361 & 0.014316 & 0.01677 & 0.014316 & 0.016226 & 0.012625 & 0.005551 & 0.002378 \\
 & RF & 0.015366 & 0.013036 & 0.015366 & 0.015552 & 0.021972 & 0.013453 & 0.217082 & 0.019841 & 0.034433 & 0.019841 & 0.021235 & 0.024199 & 0.011429 & 0.066725 \\
 & LR & 0.010999 & 0.012312 & 0.010999 & 0.012642 & 0.017529 & 0.008787 & 20.67326 & 0.008318 & 0.027902 & 0.008318 & 0.016299 & 0.017718 & 0.0051 & 0.001585 \\
 & DT & 0.012319 & 0.015275 & 0.012319 & 0.013563 & 0.020260 & 0.014356 & 0.346420 & 0.026952 & 0.029129 & 0.026952 & 0.028099 & 0.031742 & 0.018038 & 0.007679 \\
  \cmidrule{2-9} \cmidrule{10-16}
\multirow{7}{1.5cm}{Auto-Encoder}
  & SVM & 0.01077 & 0.01306 & 0.01077 & 0.01332 & 0.00125 & 0.00060 & 0.10881 & 0.00920 & 0.01191 & 0.00920 & 0.01139 & 0.00167 & 0.00084 & 0.41194 \\
 & NB & 0.02602 & 0.02366 & 0.02602 & 0.01856 & 0.01737 & 0.00400 & 0.01645 & 0.12364 & 0.07618 & 0.12364 & 0.09485 & 0.03558 & 0.01111 & 0.04840 \\
 & MLP & 0.01108 & 0.01226 & 0.01108 & 0.01075 & 0.01659 & 0.00863 & 1.02989 & 0.01107 & 0.01326 & 0.01107 & 0.01200 & 0.01462 & 0.00857 & 0.74852 \\
 & KNN & 0.01390 & 0.01360 & 0.01390 & 0.01388 & 0.01856 & 0.01095 & 0.01898 & 0.01132 & 0.01190 & 0.01132 & 0.01157 & 0.01692 & 0.01040 & 0.02262 \\
 & RF & 0.00887 & 0.00789 & 0.00887 & 0.01015 & 0.01636 & 0.01034 & 0.55832 & 0.00792 & 0.01075 & 0.00792 & 0.00992 & 0.01423 & 0.00511 & 0.36076 \\
 & LR & 0.01172 & 0.01354 & 0.01172 & 0.01413 & 0.00142 & 0.00065 & 0.13987 & 0.00877 & 0.01212 & 0.00877 & 0.01124 & 0.00138 & 0.00081 & 0.30021 \\
 & DT & 0.00682 & 0.01036 & 0.00682 & 0.00842 & 0.01239 & 0.00804 & 0.09799 & 0.00890 & 0.01560 & 0.00890 & 0.01162 & 0.02010 & 0.00778 & 0.21743 \\
  \cmidrule{2-9} \cmidrule{10-16}
\multirow{7}{1.5cm}{SeqVec}
 & SVM & 0.01041 & 0.00989 & 0.01041 & 0.01141 & 0.00206 & 0.00142 & 0.82510 & 0.00921 & 0.00812 & 0.00921 & 0.00927 & 0.00116 & 0.00061 & 0.25855 \\
 & NB & 0.04815 & 0.02762 & 0.04815 & 0.03957 & 0.02215 & 0.01374 & 0.09680 & 0.10878 & 0.04510 & 0.10878 & 0.07567 & 0.03016 & 0.01570 & 0.04303 \\
 & MLP & 0.00536 & 0.00836 & 0.00536 & 0.00570 & 0.01112 & 0.00884 & 1.48493 & 0.00943 & 0.01155 & 0.00943 & 0.01044 & 0.02354 & 0.01569 & 0.77613 \\
 & KNN & 0.00593 & 0.00899 & 0.00593 & 0.00760 & 0.01623 & 0.00816 & 0.04467 & 0.01221 & 0.01064 & 0.01221 & 0.00991 & 0.01766 & 0.00902 & 0.02768 \\
 & RF & 0.00486 & 0.00633 & 0.00486 & 0.00462 & 0.00863 & 0.00788 & 0.79517 & 0.00694 & 0.01022 & 0.00694 & 0.00697 & 0.01370 & 0.00901 & 0.33069 \\
 & LR & 0.00997 & 0.00953 & 0.00997 & 0.01077 & 0.00184 & 0.00110 & 0.35018 & 0.01052 & 0.00901 & 0.01052 & 0.01049 & 0.00148 & 0.00071 & 0.10309 \\
 & DT & 0.00474 & 0.00743 & 0.00474 & 0.00629 & 0.01299 & 0.00892 & 0.66687 & 0.00974 & 0.00994 & 0.00974 & 0.00919 & 0.02002 & 0.01160 & 0.07449 \\
 \cmidrule{2-9} \cmidrule{10-16}
 \multirow{7}{1.5cm}{ESM-2} & 
 SVM & 0.009697 & 0.009291 & 0.009697 & 0.010528 & 0.000403 & 0.000000 & 1.319865  & 0.003646 & 0.003405 & 0.003646 & 0.0036 & 0.006134 & 0.004234 & 0.621203 \\
 & NB & 0.000398 & 0.002004 & 0.000398 & 0.000983 & 0.000934 & 0.000501 & 0.181666 & 0.003388 & 0.004745 & 0.003388 & 0.003222 & 0.00557 & 0.005152 & 0.020712 \\
 & MLP & 0.009697 & 0.009291 & 0.009697 & 0.010528 & 0.000403 & 0.000000 & 0.906801 & 0.005089 & 0.005203 & 0.005089 & 0.005084 & 0.006778 & 0.002291 & 0.973078 \\
 & KNN & 0.009697 & 0.009280 & 0.009697 & 0.010518 & 0.000402 & 0.000004 & 0.094752 & 0.002152 & 0.002061 & 0.002152 & 0.002167 & 0.005009 & 0.004374 & 0.023182 \\
 & RF & 0.009697 & 0.009291 & 0.009697 & 0.010528 & 0.000403 & 0.000000 & 0.309601 & 0.002641 & 0.002657 & 0.002641 & 0.002687 & 0.005501 & 0.005179 & 1.211423 \\
 & LR & 0.009697 & 0.009291 & 0.009697 & 0.010528 & 0.000403 & 0.000000 & 1.165850 & 0.008082 & 0.007644 & 0.008082 & 0.007976 & 0.009914 & 0.007188 & 0.605703 \\
 & DT & 0.009697 & 0.009291 & 0.009697 & 0.010528 & 0.000403 & 0.000000 & 0.002509 & 0.002489 & 0.002241 & 0.002489 & 0.002377 & 0.005273 & 0.004351 & 0.474218 \\
\cmidrule{2-9} \cmidrule{10-16}
\multirow{7}{1.5cm}{TAPE}
 & SVM & 0.010316 & 0.012481 & 0.010316 & 0.012233 & 0.024412 & 0.009227 & 0.349132 & 0.003124 & 0.003241 & 0.003124 & 0.003518 & 0.016423 & 0.007611 & 0.574466 \\
 & NB & 0.159849 & 0.02857 & 0.159849 & 0.138144 & 0.026035 & 0.019692 & 0.061332 & 0.010964 & 0.014088 & 0.010964 & 0.007802 & 0.005716 & 0.003786 & 0.050925 \\
 & MLP & 0.007817 & 0.008321 & 0.007817 & 0.007474 & 0.011352 & 0.002786 & 1.10145 & 0.010378 & 0.006162 & 0.010378 & 0.008163 & 0.016417 & 0.008435 & 0.436171 \\
 & KNN & 0.017575 & 0.010538 & 0.017575 & 0.016839 & 0.012342 & 0.001747 & 0.005465 & 0.007472 & 0.010078 & 0.007472 & 0.00745 & 0.006157 & 0.002939 & 0.008291 \\
 & RF & 0.007281 & 0.005008 & 0.007281 & 0.008343 & 0.016176 & 0.004181 & 0.23628 & 0.005753 & 0.006526 & 0.005753 & 0.006462 & 0.004503 & 0.001946 & 0.614082 \\
 & LR & 0.008777 & 0.013602 & 0.008777 & 0.009027 & 0.006907 & 0.003453 & 0.787583 & 0.011012 & 0.010556 & 0.011012 & 0.010163 & 0.018999 & 0.00845 & 1.044759 \\
 & DT & 0.011427 & 0.009725 & 0.011427 & 0.010579 & 0.018905 & 0.010748 & 0.313134 & 0.01399 & 0.011397 & 0.01399 & 0.012599 & 0.017441 & 0.008658 & 0.902225 \\

\cmidrule{2-9} \cmidrule{10-16}
\multirow{7}{1.5cm}{Weighted PSSKM Kernel (ours)}
      & SVM & 0.00648 & 0.00599 & 0.00498 & 0.00745 & 0.01347 & 0.00793 & 0.32149
       & 0.00499 & 0.00674 & 0.00612 & 0.00687 & 0.01611 & 0.00974 & 0.22764\\
 & NB & 0.01322 & 0.01478 & 0.01395 & 0.01544 & 0.01815 & 0.01133 & 0.53631
 & 0.01527 & 0.01412 & 0.01357 & 0.01314 & 0.01809 & 0.01167 & 0.51931\\
 & MLP & 0.00698 & 0.00894 & 0.00643 & 0.00845 & 0.01322 & 0.00813 & 3.13462
 & 0.00925 & 0.00814 & 0.00853 & 0.00847 & 0.01363 & 0.00874 & 1.49651\\
 & KNN & 0.01231 & 0.01056 & 0.01487 & 0.01254 & 0.01998 & 0.01676 & 0.76152
 & 0.01156 & 0.01013 & 0.01432 & 0.01216 & 0.03311 & 0.01356 & 0.54301\\
 & RF & 0.00542 & 0.00854 & 0.00322 & 0.00233 & 0.01543 & 0.00953 & 0.91258
 & 0.00567 & 0.00854 & 0.00565 & 0.00542 & 0.01653 & 0.00756 & 0.78123\\
 & LR & 0.00435 & 0.00743 & 0.00434 & 0.00643 & 0.00734 & 0.00532 & 0.26145
 & 0.00654 & 0.00432 & 0.00632 & 0.00425 & 0.00753 & 0.00594 & 0.34325\\
 & DT & 0.00543 & 0.00512 & 0.00743 & 0.00832 & 0.01843 & 0.00743 & 0.29754
  & 0.00712 & 0.00578 & 0.00713 & 0.00835 & 0.01831 & 0.00845 & 0.29543\\
         \bottomrule
         \end{tabular}
}
 \caption{Classification results (standard deviation values over $5$ runs) on \textbf{Spike7k} and \textbf{Protein Subcellular} datasets for different evaluation metrics.}
    \label{tbl_results_classification_std}
\end{table*}

\end{document}